\documentclass{article}

\usepackage[T1]{fontenc}
\usepackage{microtype}
\usepackage{graphicx}
\usepackage{subcaption}
\usepackage{booktabs} 

\usepackage{hyperref}

\usepackage[accepted]{icml2026}

\usepackage{amsmath}
\usepackage{amssymb}
\usepackage{mathtools}
\usepackage{amsthm}
\usepackage{pifont}

\newcommand{\cmark}{\ding{51}} 

\usepackage[capitalize,noabbrev]{cleveref}

\usepackage{enumitem}
\usepackage{mathtools}
\usepackage{multirow}
\usepackage{tcolorbox}
\usepackage{adjustbox}

\usepackage{amsmath,amsfonts,bm,amssymb}

\def\1{\bm{1}}

\def\vone{{\bm{1}}}

\def\vu{{\bm{u}}}

\def\vx{{\bm{x}}}
\def\vy{{\bm{y}}}

\def\mA{{\bm{A}}}
\def\mB{{\bm{B}}}
\def\mC{{\bm{C}}}

\def\mS{{\bm{S}}}

\def\mX{{\bm{X}}}
\def\mY{{\bm{Y}}}

\DeclareMathAlphabet{\mathsfit}{\encodingdefault}{\sfdefault}{m}{sl}
\SetMathAlphabet{\mathsfit}{bold}{\encodingdefault}{\sfdefault}{bx}{n}

\def\gI{{\mathcal{I}}}

\theoremstyle{definition}

\usepackage{accents}
\newlength{\dhatheight}

\DeclarePairedDelimiter{\abs}{\lvert}{\rvert}
\DeclarePairedDelimiter{\norm}{\lVert}{\rVert}

\theoremstyle{plain}

\theoremstyle{definition}

\theoremstyle{remark}

\usepackage[textsize=tiny]{todonotes}

\icmltitlerunning{Local Causal Attribution of Chain-of-Thought Reasoning}

\begin{document}

\twocolumn[
  \icmltitle{Local Causal Attribution of Chain-of-Thought Reasoning}



  \icmlsetsymbol{equal}{*}

  \begin{icmlauthorlist}
    \icmlauthor{Dennis Wei}{ibm}
    \icmlauthor{Yannis Belkhiter}{ibm}
    \icmlauthor{Erik Miehling}{ibm}
    \icmlauthor{Radu Marinescu}{ibm}
  \end{icmlauthorlist}

  \icmlaffiliation{ibm}{IBM Research}

  \icmlcorrespondingauthor{Dennis Wei}{dwei@us.ibm.com}

  \icmlkeywords{Machine Learning, ICML}

  \vskip 0.3in
]



\printAffiliationsAndNotice{}  

\begin{abstract}
Understanding the causal structure of a language model's thought process is a problem of significant importance for both transparency and safety. In this work, we take a \emph{local} approach toward this goal by analyzing the causal relationships among individual components, termed units, of a given, \emph{specific} chain-of-thought trace. We construct a structural causal model on these units and relate each unit to the log probability of generating (subsequent) output units. Our algorithm, termed AttriCoT, is a black-box method that performs attribution by estimating importance parameters in the structural causal model using $O(U)$ forward passes through the model, where $U$ is the number of units. Evaluation of perturbation curves across 5 datasets and 4 reasoning models shows that AttriCoT produces attributions that are more faithful to the model's behavior than alternative methods. The attribution results also reveal notable differences in thought structure between models and domains.
\end{abstract}

\section{Introduction}

The success of LLMs and large reasoning models (LRMs) in performing reasoning is due in large part to their generation of long sequences of reasoning steps known as chains-of-thought (CoTs). 
Moreover, because of the readability of their text, CoTs appear to promise the additional benefit of greater insight into the model's reasoning process. 
It is perhaps paradoxical then that the introduction of CoTs has raised more questions than it has answered. 
Prominent among these questions is that of \emph{CoT faithfulness}, which refers to how well the verbalized CoT text represents the model's underlying (latent) thought process. 
Research has shown that CoT text may disregard manipulation of the model's input or injection of helpful hints, even constructing elaborate but false justifications when given incorrect hints \cite{turpin2023languagemodelsdontsay, chen2025reasoning, arcuschin2025chain}, with evidence that larger models may rely \emph{less} on their stated reasoning than smaller models \cite{lanham2023measuring}. On the other hand, more recent work \cite{vonarx2025cot} illustrates that CoTs may become more faithful for more difficult tasks.

Beyond the debate about faithfulness of CoT \emph{text}, there exist other questions about the \emph{mechanisms} that underlie CoTs. 
Recent work on CoT faithfulness \cite{barez2025chain,zaman2025chainofthoughtreallyexplainabilitychainofthought} has also moved toward this direction, arguing that progress in faithfulness can be achieved via \emph{causal} CoTs that demonstrably affect the model's final answer. 
These questions, about mechanisms and improving faithfulness, motivate  
investigation of the causal structure within a CoT trace. The focus of study thus becomes analyzing the impact of earlier thoughts/steps on later ones. Understanding this structure can provide insights into why certain steps are produced, whether intermediate reasoning is consequential, and where/when unfaithfulness may arise. 

Causal structure within a CoT can be investigated at different levels. First, there is the level of the model's \emph{internal} reasoning, for example the mechanisms by which the internal representation of a thought affects that of a later thought, which we do not tackle in this work. Then there is the level of causal effects on \emph{distributions} over reasoning steps or trajectories. This level involves resampling of the CoT and was explored in recent work \cite{macar2025thoughtbranchesinterpretingllm,bogdan2025thoughtanchorsllmreasoning}. Our paper focuses on a third level that we call \emph{local causality}, where ``local'' refers to restriction to a single, specific CoT. This notion of local causality is useful for explaining specific CoT steps (that a user might see), rather than a distribution over steps. Moreover, it offers a more computationally efficient approach to understanding CoT structure, requiring only forward passes through the model as opposed to the resampling of \citet{macar2025thoughtbranchesinterpretingllm,bogdan2025thoughtanchorsllmreasoning}, which is significantly more expensive.

We develop a black-box method, termed \emph{AttriCoT}, for estimating causal relationships at the local level, within a given CoT. We generalize the concept of a CoT step to that of a \emph{unit}, i.e., user-configurable segments of the output from the model as well as the input to it. For each unit in the model output, {AttriCoT} estimates the effects that prior units have on the log probability of generating the unit of interest, thereby attributing it to prior units. AttriCoT works by 1) performing interventions on units, 2) computing forward passes to determine the log probability effects of interventions, and 3) fitting the observed effects to a linear structural equation model. Computational tractability is maintained by restricting the number of forward passes to scale linearly with the number of units. The result can be represented as a matrix of pairwise importance scores, where element $(i, j)$ represents influence of unit~$i$ on unit~$j$. Such characterization of causal effects between all pairs of CoT steps has been less studied to date. 

Our focus on locality aligns with and extends a well-established line of work on \emph{local explanations}, which aim to explain specific outputs of ML models on specific inputs, ranging from predictions of classifiers~\cite{ribeiro2016why,lundberg2017shap} to responses of LLMs~\cite{sarti-etal-2023-inseq,miglani2023using,enouen-etal-2024-textgenshap,cohen-wang2024contextcite,monteiro-paes-etal-2025-mexgen}. We extend local explanations to a setting where there is no clear distinction between inputs and outputs, since LLMs consume their generated CoTs as input. 

We evaluate AttriCoT against two types of baselines, namely the sentence-to-sentence masking method of {Thought Anchors} \cite{bogdan2025thoughtanchorsllmreasoning} and prompting, on CoTs of 4 LRMs across 5 datasets. Our results show that AttriCoT consistently outperforms these baselines in terms of behavioral faithfulness to the LRM that generated the CoT (as measured by the area under a perturbation curve, AUPC). For instance, on GSM8K, AttriCoT achieves up to 15-30\% and 70-165\% AUPC improvement relative to Thought Anchors and prompting baselines across models, respectively. 
Furthermore, we analyze the importance score matrices produced by AttriCoT by computing metrics that summarize the matrix along different dimensions. These metrics can reveal intriguing differences in CoT structure between models and datasets, and we believe that this manner of analysis holds significant potential.

Our \textbf{contributions} can be summarized as follows:
\begin{enumerate}[nosep,left=5pt]
    \item [(\textbf{1})] We propose {AttriCoT}, a black-box local explanation method for attributing CoT steps and other output units (e.g., of the answer) to prior units, based on a structural causal model and using a linear number of forward passes. 
    \item [(\textbf{2})] We show that AttriCoT's attributions are consistently more faithful to the model's behavior than alternative methods. 
    \item [(\textbf{3})] We show that metrics computed from AttriCoT's attribution matrices can open the door toward greater insights into CoTs.
\end{enumerate}

\section{Related Work}
\label{sec:related_work}

\paragraph{CoT attribution methods.} Various methods have recently been proposed for the attribution analysis of CoTs. To frame our comparison, we position related work along three axes: causality, granularity, and access level (see Table \ref{tab:related-works-v2} in Appendix \ref{sec:appendix-related-work} for a detailed comparison). Broadly, perturbation-based methods \cite{turpin2023languagemodelsdontsay, atanasova2023faithfulnesstestsnaturallanguage, parcalabescu2024measuringfaithfulnessselfconsistencynatural} measure correlations with respect to the final output, but do not establish causality. White-box causal methods \cite{zhao2026verifyingchainofthoughtreasoningcomputational, ye2026mechanisticevidencefaithfulnessdecay} aim to reveal causal structures but require access to the model's internals. Black-box step-level causal methods \cite{paul2024makingreasoningmattermeasuring, macar2025thoughtbranchesinterpretingllm} are the most relevant category to our work. They perturb individual steps (without access to the model's internals) and measure the attribution as the effect on the final answer. This does have limitations, however, in that the impacts of intermediate steps are not accounted for. 

\paragraph{Beyond answer-level attribution.} To the best of our knowledge, Thought Anchors, and specifically their sentence-to-sentence masking method \citep[Sec.~5, App.~M]{bogdan2025thoughtanchorsllmreasoning}, is the only work that moves beyond answer-level attribution. Thought Anchors attributes every unit in the CoT to prior ones by masking each sentence and performing a forward pass to determine the effects. It measures changes in token distributions using Kullback-Leibler (KL) divergence at every token position in a target CoT unit. In contrast, AttriCoT measures the change in log probability of the given generated unit, which is fully local in the sense discussed in the introduction. Additionally, AttriCoT fits a structural causal model and can thus exploit information from any intervention type.

\section{A Structural Causal Model for Chain-of-Thought}
\label{sec:causal_model}

In this section, we present a structural causal model (SCM) for the steps in an LRM's CoT, allowing us to define the causal effect of one CoT step on another and thereby attribute steps to earlier steps.

Let $\vx$ be a given input, e.g., prompt, to the LRM and $\vy$ be the corresponding output, e.g., solution, where $\vy$ consists of both the LRM's CoT and the final answer. We assume that $\vy$ has been split into what we generally call \emph{units}, $(\vy_1, \dots, \vy_T)$. These include steps or ``thoughts'' in the CoT, followed by the final answer, which may also be split into multiple units (our notation does not distinguish between CoT units and answer units). We consider the input $\vx$ to be split into units $(\vx_1, \dots, \vx_S)$ as well. For example, $\vx$ may be a question and $(\vx_1, \dots, \vx_S)$ may be sentences in the question that give different pieces of information. We will only consider attributing output units $\vy_t$ to other units, not input units $\vx_s$ (i.e., we will not model their causes). 
We use the symbols $\vx_s$, $\vy_t$ to refer to units at a more abstract level as well as the sequences of tokens that they comprise.

In the classical setting of an SCM with scalar variables $(x_1, \dots, x_S, y_1, \dots, y_T)$, the model is specified by structural equations, one for each variable \cite{pearl2009causality}. For example, the structural equation for $y_t$ 
takes the general form 
\begin{equation}\label{eq:sem_scalar}
    y_t = F_t(\vx, \vy_{< t}, \vu),
\end{equation}
where we assume an auto-regressive property that mirrors how LLMs/LRMs operate, $\vy_{< t} = (\vy_1, \dots, \vy_{t-1})$ is the set of output units that causally precede $y_t$, and $\vu$ represents exogenous variables.
In the LRM setting however, all variables are natural language units, which have representations as sequences of tokens.  
In particular, we now have such a sequence $\vy_t$ on the left side of \eqref{eq:sem_scalar}. We would like $F_t$ to remain a scalar, real-valued function to quantify the influence that earlier units have on $\vy_t$. 
We propose therefore to ``scalarize'' the unit $\vy_t$ with a \emph{scalarizer} function $S_t$, yielding 
\begin{equation}\label{eq:sem_scalarized}
    S_t(\vy_t) = F_t(\vx, \vy_{< t}, \vu).
\end{equation}
We choose $S_t$ to be the log probability of the token sequence of $\vy_t$ conditioned on previous units:
\begin{align}\label{eq:sem_logprob}
    S_t(\vy_t) &= \log P(\vy_t \mid \vx, \vy_{< t})\nonumber\\
    &= \sum_{j=1}^{\abs{\vy_t}} \log P(y_{t,j} \mid \vx, \vy_{< t}, \vy_{t,<j}), \quad t = 1,\dots,T,
\end{align}
where we have used the chain rule of probability to express the log probability as a sum of log probabilities of tokens $y_{t,j}$ in $\mathbf{y}_t$. 
The fact that $S_t(\vy_t)$ is a function of the given, generated unit $\vy_t$ reflects our desire for locality, as discussed in the introduction. The use of log probability \eqref{eq:sem_logprob} allows causal effects to be measured more efficiently, using non-generative forward passes only. 

For the function $F_t$, we propose that it be an interpretable model, in line with our goal of providing local explanations. In this work, we use a linear model with ``interpretable features'' that indicate whether units are present or absent (replaced by the empty sequence). Let $\mathbb{I}(\vx)$ denote such an indicator function (equals $1$ if unit $\vx$ is present and $0$ if absent). The linear model is given by
\begin{multline}\label{eq:linear_model}
    F_t(\vx, \vy_{< t}, \vu) = \sum_{s=1}^S \alpha_{s,t} \mathbb{I}(\vx_s) + \sum_{\tau=1}^{t-1} \beta_{\tau,t} \mathbb{I}(\vy_{\tau}) + \gamma_t + u_t, \\ t = 1,\dots,T,
\end{multline}
with coefficients $\alpha_{s,t}$ and $\beta_{\tau,t}$ and intercept $\gamma_t$. Other interpretable models could also be used, for example those involving Boolean functions of $\mathbb{I}(\vx_s), \mathbb{I}(\vy_\tau)$.

Equations \eqref{eq:sem_scalarized}--\eqref{eq:linear_model} together attribute the conditional log probability of output unit $\vy_t$ to the presence or absence of prior units. We therefore interpret coefficients $\alpha_{s,t}$ and $\beta_{\tau,t}$ as importance scores for generating $\vy_t$. (The intercept $\gamma_t$ also has an interpretation as the log probability of $\vy_t$ with an empty prefix.) For future use, we collect all $\alpha_{s,t}$ into a matrix $\mA \in \mathbb{R}^{S\times T}$. Simlarly, we collect $\beta_{\tau,t}$ into a \emph{strictly upper-triangular} matrix $\mB \in \mathbb{R}^{T\times T}$, i.e., $\beta_{\tau,t} = 0$ for $\tau \geq t$. The upper-triangular property of $\mB$ is the manifestation of the auto-regressive property of LRMs for the case of the linear structural equations in \eqref{eq:linear_model}.

\begin{figure*}[ht]
    \centering
    \includegraphics[width=0.8\linewidth]{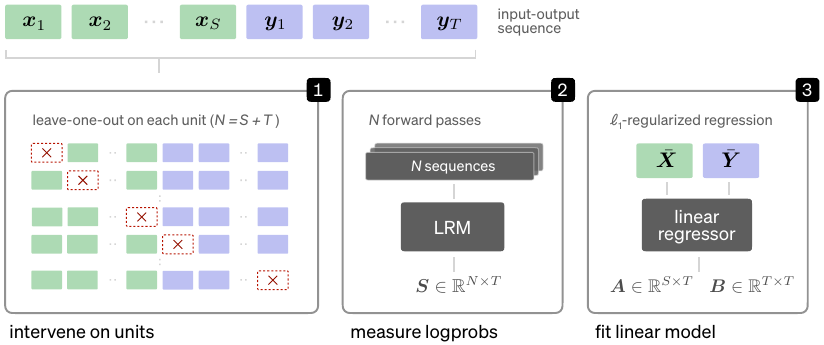}
    \caption{The AttriCoT method consists of three main steps: performing unit-level interventions (\raisebox{-0.15em}{\includegraphics[height=0.9em]{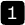}}), measuring the impact of the interventions on log probabilities (\raisebox{-0.15em}{\includegraphics[height=0.9em]{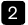}}), and estimating the attribution scores by fitting a linear model (\raisebox{-0.15em}{\includegraphics[height=0.9em]{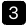}}).}
    \label{fig:attricot}
\end{figure*}

\section{CoT Attribution Methods}
\label{sec:method}

In Section~\ref{sec:method:attricot}, we describe our proposed AttriCoT method for attributing CoT steps and other output units to prior units. Section~\ref{sec:method:white-box} briefly describes an exploratory white-box method to reduce computational cost.

\subsection{AttriCoT}
\label{sec:method:attricot}
AttriCoT performs CoT attribution by fitting the structural equations in \eqref{eq:linear_model} via linear regression, thereby estimating the local causal effects of prior units. 
The process involves three steps: 1) Selecting and performing interventions on prior units, which corresponds to defining the feature matrix for the linear regressions; 2) Measuring the log probabilities of output units that result from interventions, which corresponds to obtaining the target values; and 3) Actually performing the linear regressions. These steps are described in more detail below.

\paragraph{\raisebox{-0.1em}{\includegraphics[height=0.9em]{assets/tag1.pdf}} Sample and perform interventions.} 
Given an input-output sequence of units $(\vx, \vy) = (\vx_1, \dots, \vx_S, \vy_1, \dots, \vy_T)$, we perform interventions on units to determine their causal effects. We denote by $(\vx^{(i)}, \vy^{(i)}), i = 1,\dots,N$ the perturbed sequences resulting from intervention, and also define $(\vx^{(0)}, \vy^{(0)}) = (\vx, \vy)$ to be the original sequence for $i = 0$. This intervention step involves making three choices: 1) the number of interventions $N$, 2) which units to intervene on, and 3) how to intervene. For 3), AttriCoT simply removes the selected units from the sequence, consistent with setting the indicator functions in \eqref{eq:linear_model} to zero. For 1), we restrict $N$ to be a multiple of the total number of units $S + T$. As described further in Step 2, $N$ is also the number of forward passes through the LRM, so choosing $N = O(S + T)$ helps to control this computational cost. 

A basic intervention strategy with $N = S + T$, and the default for AttriCoT, is the \emph{leave-one-out} (LOO) procedure, which removes each unit $\vx_s$ or $\vy_t$ one at a time. This yields perturbed sequences 
\begin{align}
    &\big(\vx^{(i)}, \vy^{(i)}\bigr)\nonumber\\ 
    &= \begin{cases}
        \left(\vx_1, \dots, \vx_{i-1}, \vx_{i+1}, \dots, \vx_S, \vy_1, \dots, \vy_T\right),\\ 
        \qquad \qquad \qquad \qquad \qquad \qquad i = 1,\dots,S,\\
        \left(\vx_1, \dots, \vx_S, \vy_1, \dots, \vy_{i-S-1}, \vy_{i-S+1}, \dots, \vy_T\right),\\ 
        \qquad \qquad \qquad \qquad \qquad \qquad i = S+1,\dots,S+T.
    \end{cases}
\end{align}
When $N > S + T$, we also have interventions in which more than one unit is intervened on simultaneously, but we limit the number of such units to a small integer. Please see Appendix~\ref{sec:expt:methods_addl} for more details.

\paragraph{\raisebox{-0.1em}{\includegraphics[height=0.9em]{assets/tag2.pdf}} Measure log probabilities.}
The second step in AttriCoT is to measure the scalarized effects on downstream units (i.e., the changes in their log probabilities) of the interventions of the first step. Each perturbed sequence $(\vx^{(i)}, \vy^{(i)})$ is passed as input to the LRM and a forward pass is performed to compute the conditional log probabilities of all tokens, $\log P(y_{t,j}^{(i)} \mid \vx^{(i)}, \vy_{< t}^{(i)}, \vy_{t,<j}^{(i)})$. For the original sequence $i = 0$, these log probabilities are either pre-computed and given as input to AttriCoT, or we perform one additional forward pass on $(\vx^{(0)}, \vy^{(0)})$. The log probabilities are then aggregated over the tokens in each output unit $\vy_t$, where instead of the sum in \eqref{eq:sem_logprob}, we take the mean to normalize by the length of the unit. The result is 
\begin{multline}\label{eq:unit_log_probs}
    S_t\left(\vy_t^{(i)}\right) 
    = \frac{1}{\abs[\big]{\vy_t^{(i)}}} \sum_{j=1}^{\abs{\vy_t^{(i)}}} \log P\left( y_{t,j}^{(i)} \mid \vx^{(i)}, \vy_{< t}^{(i)}, \vy_{t,<j}^{(i)} \right), \\ i = 0,\dots,N, \; t = 1,\dots, T.
\end{multline}
The above computation requires $O(N)$ forward passes for $N$ interventions, as well as access to output log probabilities or logits. To increase computational efficiency, these forward passes are batched as much as possible. It is worth noting that AttriCoT is able to attribute all $T$ output units using $O(N) = O(S + T)$ forward passes, i.e., the inference cost is linear in the number of units (not quadratic).

\paragraph{\raisebox{-0.1em}{\includegraphics[height=0.9em]{assets/tag3.pdf}} Fit linear model.}
The third step is to perform linear regression to estimate the coefficients $\alpha_{s,t}$ and $\beta_{\tau,t}$ in \eqref{eq:linear_model}, using the indicator function feature values from the first step and the log probability target values from the second step. 
To specify the objective function for these linear regressions, we first collect the indicator values for input units into a feature matrix $\bar{\mX} \in \mathbb{R}^{(N+1)\times S}$ with entries $\bar{x}_{i,s} = \mathbb{I}(\vx_s^{(i)})$ (including $i=0$ which yields an all-ones row), and those for output units into another matrix $\bar{\mY} \in \mathbb{R}^{(N+1)\times T}$ with entries $\bar{y}_{i,t} = \mathbb{I}(\vy_t^{(i)})$. We similarly define a target matrix $\mS \in \mathbb{R}^{(N+1)\times T}$ via $S_{i,t} = S_t(\vy_t^{(i)})$ from \eqref{eq:unit_log_probs}. 
Then for output unit $\vy_t$, we regress the $t$th target column $\mS_{:,t}$ on $\bar{\mX}$ and $\bar{\mY}$ to obtain the $t$th columns of $\mA$ and $\mB$. However, we also need to exclude rows (interventions) $i$ where $\vy_t$ itself was removed, since it is not possible to compute the log probability of $\vy_t$ in those cases. Let $\gI = \{i: \bar{y}_{i,t} = 1\}$ be the set of rows where $\vy_t$ is preserved. The final form of the regression problem is therefore
\begin{multline}
    \min_{\mA_{:,t}, \mB_{<t,t}, \gamma_t} \; \norm*{\bar{\mX}_{\gI,:} \mA_{:,t} + \bar{\mY}_{\gI,<t} \mB_{<t,t} + \gamma_t \vone - \mS_{\gI,t}}_2^2 \\ + \lambda \left(\norm{\mA_{:,t}}_1 + \norm{\mB_{<t,t}}_1\right),
\end{multline}
where we use the standard least-squares loss with optional $\ell_1$ regularization.

\subsection{Exploratory White-Box Method based on Attention Intervention}
\label{sec:method:white-box}

We explored the possibility of further reducing the computational cost of AttriCoT (dominated by the $O(N)$ forward passes through the model) in the case of white-box access to the model. When AttriCoT removes a unit, it removes the unit's tokens from the token sequence that is input to the model and performs a full forward pass to compute log probabilities. For the white-box approach, we instead considered intervening on the attention weights of the unit's tokens in a single Transformer block of the model. The outputs of the attention heads in the block are then recomputed and propagated to the model's logits through a ``shallow network.'' This approach is significantly less expensive because it requires only one full forward pass to obtain the original attention weights and other quantities needed to perform the interventions. Subsequently, $N$ shallow passes suffice to compute perturbed attention outputs and propagate them through the shallow network to the logits.

Appendix~\ref{sec:attention_intervention} provides more details on this attention intervention method. In brief, we intervene by setting attention weights to zero and show how the attention output is recomputed. The shallow network consists essentially of an MLP layer and the language modeling head of the LLM, and there are choices of which Transformer block to intervene on and how to construct the shallow network. We experimented with two variants, an ``untrained'' one in which the shallow network is constructed from modules of the LLM, and a ``trained'' one where the shallow network is then fine-tuned.

\section{Evaluation of Attribution Quality}
\label{sec:expt}

In this section, we present experiments that evaluate the quality of AttriCoT's attributions in comparison to baselines.

\subsection{Experimental Setup}
\label{sec:expt:setup}

\paragraph{Datasets.}
We selected five 
datasets for the evaluation. First, we attributed reasoning traces for mathematical reasoning tasks. We selected two complementary datasets, namely GSM8K \cite{cobbe2021gsm8k} and MATH500 \cite{hendrycks2021measuringmathematicalproblemsolving}. GSM8K consists of high-school-level mathematical questions while MATH500 includes more challenging problems. Next, we evaluated the attribution methods on general reasoning and knowledge tasks. To this end, we included GPQA-Diamond \cite{rein2023gpqagraduatelevelgoogleproofqa}, including PhD-level questions in physics and biology, as well as MMLU-Pro \cite{wang2024mmluprorobustchallengingmultitask}, including 14 different domains. We also included a third task of reasoning on logic problems, represented by the ZebraLogic \cite{zebralogic2025} benchmark. Table \ref{tab:reasoning-datasets} in Appendix \ref{sec:appendix-experiment-design} further describes the datasets. 

\textbf{Models.~}
To generate reasoning traces on which to perform attribution, we selected four state-of-the-art \emph{open-source} reasoning models. First, we selected \texttt{DS-Qwen3-8B} \cite{deepseekai2025deepseekr1incentivizingreasoningcapability,qwen3technicalreport} for its good performance given its model size. Second, we selected \texttt{DS-Llama-8B} \cite{deepseekai2025deepseekr1incentivizingreasoningcapability,grattafiori2024llama3herdmodels} for being of the same size as the first one, but distilled on a different model architecture. We also selected \texttt{DS-Qwen-14B} \cite{deepseekai2025deepseekr1incentivizingreasoningcapability}, a larger model, and \texttt{Qwen3-8B} \cite{qwen3technicalreport}, as the non-DeepSeek-distilled counterpart to \texttt{DS-Qwen3-8B}. Table \ref{tab:selected-models} in Appendix \ref{sec:appendix-experiment-design} gives details on the models that we selected.

\begin{table*}[t]
\centering
\caption{AUPC ($\uparrow$) of all attribution methods on the GSM8K dataset.}
\label{tab:aupc_gsm8k}
\small
\begin{tabular}{lcccc}
\toprule
Algorithm & DS-Llama-8B & DS-Qwen-14B & DS-Qwen3-8B & Qwen3-8B \\
\midrule
AttriCoT & \textbf{231.00} $\pm$ 2.17 & \textbf{216.70} $\pm$ 1.79 & \textbf{138.64} $\pm$ 1.05 & \textbf{166.59} $\pm$ 1.53 \\
AttriCoT-2x & \underline{232.39} $\pm$ 2.13 & \underline{218.55} $\pm$ 1.80 & \underline{139.73} $\pm$ 1.05 & \underline{168.03} $\pm$ 1.54 \\
TA-KL & 183.29 $\pm$ 1.75 & 169.47 $\pm$ 1.49 & 116.05 $\pm$ 1.12 & 98.53 $\pm$ 1.03 \\
TA-KL (no log) & 195.61 $\pm$ 2.19 & 181.82 $\pm$ 1.71 & 123.42 $\pm$ 1.20 & 161.28 $\pm$ 1.60 \\
Prompted (self) & 88.65 $\pm$ 1.49 & 95.52 $\pm$ 1.36 & 61.39 $\pm$ 0.67 & 67.45 $\pm$ 0.98 \\
Prompted-ICL (self) & 107.29 $\pm$ 1.79 & 117.77 $\pm$ 1.60 & 60.38 $\pm$ 0.65 & 50.49 $\pm$ 0.71 \\
Prompted (counterpart) & --- & 147.27 $\pm$ 1.50 & 60.20 $\pm$ 0.71 & --- \\
Prompted-ICL (counterpart) & --- & 153.51 $\pm$ 1.35 & 58.39 $\pm$ 0.67 & --- \\
Prompted (GPT-OSS) & 96.23 $\pm$ 1.30 & 94.87 $\pm$ 1.28 & 49.65 $\pm$ 0.60 & 54.78 $\pm$ 0.77 \\
Prompted-ICL (GPT-OSS) & 135.79 $\pm$ 1.47 & 142.58 $\pm$ 1.46 & 62.06 $\pm$ 0.73 & 72.06 $\pm$ 0.92 \\
Attention-based & 107.06 $\pm$ 1.25 & 87.36 $\pm$ 1.07 & 65.72 $\pm$ 0.61 & 58.31 $\pm$ 0.73 \\
Attention-based (trained) & 101.43 $\pm$ 1.35 & 111.39 $\pm$ 1.19 & 61.88 $\pm$ 0.58 & 56.02 $\pm$ 0.71 \\
\bottomrule
\end{tabular}
\end{table*}

\textbf{Attribution methods.~} We consider four categories of attribution methods: our AttriCoT method, Thought Anchors, prompting baselines, and the exploratory attention-based method of Section~\ref{sec:method:white-box}.

\emph{AttriCoT.} We evaluate two variants of AttriCoT, the LOO variant described in Section~\ref{sec:method} that uses $N = S + T$ interventions, and a ``2x'' variant that uses $N = 2(S + T)$ interventions. For the latter AttriCoT-2x variant, $S + T$ of its interventions are the same LOO ones as AttriCoT-LOO, while the other $S + T$ are ``leave-two-out,'' sampled uniformly from all pairs of units. We regard AttriCoT-LOO as our primary method and AttriCoT-2x as an enhanced variant that uses twice the computational cost. The motivation for AttriCoT-2x is that since AttriCoT works by fitting a linear model, it could potentially take advantage of additional interventions (i.e., observations) if available. Evaluating AttriCoT-2x quantifies this advantage. 

\emph{Thought Anchors.} We compare AttriCoT with the sentence-to-sentence masking method of Thought Anchors \cite{bogdan2025thoughtanchorsllmreasoning} described in Section~\ref{sec:related_work} since it 
is the closest existing method to ours. Since this method uses KL divergence to quantify causal effects, we refer to it as Thought Anchors-KL. We evaluate two versions, one that applies a log transform to the KL divergence (as specified in their paper), 
and one that does not (as implemented in their code). Additional discussion of how we adapted and improved Thought Anchors-KL can be found in Appendix~\ref{sec:expt:methods_addl}. The computational complexity of Thought Anchors-KL is similar to that of AttriCoT-LOO, requiring $S + T$ forward passes. However, unlike AttriCoT, Thought Anchors-KL is not able to exploit additional forward passes if available.

\emph{Prompting.} We also prompted LLMs to perform CoT attribution. Specifically, since our method of evaluating CoT attributions (described below) depends only on how an attribution method ranks units, we prompted the LLM to rank units in order of importance for generating each output unit, as opposed to the harder task of assigning numerical scores. To make the task easier still, we followed \cite{monteiro-paes-etal-2025-mexgen} in numbering units with tags and asking only for a ranked list of tags. Please see Appendix~\ref{sec:expt:methods_addl} for the prompt that we used and other details. Since we prompted the LLM separately for each output unit, the complexity is also $O(S + T)$, but these are generative calls which are more expensive than the non-generative forward passes of AttriCoT and Thought Anchors-KL.

We experimented with prompting different LLMs: 1) The same model that generated the CoT (i.e., self-attribution); 2) If available, a ``counterpart'' model with the same architecture that is instruct-tuned and not reasoning-distilled, specifically Qwen2.5-14B-Instruct for DS-Qwen-14B and Qwen3-8B for DS-Qwen3-8B; 3) GPT-OSS-120B, a much larger and more powerful open-weights LLM. We also experimented with providing an in-context learning (ICL) example or not.

\emph{Attention Intervention.} We evaluated the attention intervention method of Section~\ref{sec:method:white-box} only on the GSM8K dataset, both untrained and trained versions as well as intervening on different Transformer blocks. The results reported in this section are for the best Transformer block found; results for all choices of blocks can be found in Appendix~\ref{sec:expt:results_addl:white-box}.

\textbf{Evaluation method.~} We evaluate CoT attributions in terms of how behaviorally faithful they are to the model that generated the CoT, specifically by computing perturbation curves as is standard in the explainable AI/NLP literature \cite{chen-etal-2020-generating-hierarchical,ju-etal-2023-hierarchical,monteiro-paes-etal-2025-mexgen}. Given an attribution method's ranking of prior units for an output unit, the procedure is to perturb (i.e., remove) the top $k$ units, compute the resulting decrease in log probability of the output unit, and plot the curve of these log probability decreases as a function of $k$. We vary $k$ from $0$ up to $20\%$ of the number of prior units. For each sample, we average the perturbation curves corresponding to different output units (please see Appendix~\ref{sec:expt:perturb_curve_addl} for how). In this section, we summarize 
perturbation curves by computing the area under them (AUPC), where higher values are better. The full perturbation curves are deferred to Appendix~\ref{sec:expt:results_addl:perturbation}.

\subsection{Results}
\label{sec:expt:results}

We first discuss results on the GSM8K dataset, which we used as an exploratory dataset to evaluate a larger number of variations of the baselines. We then carried forward the stronger variations to the other datasets. These variations include both versions of Thought Anchors-KL (with and without log transform) and prompting the three types of LLMs (CoT model itself, non-reasoning-distilled counterpart if available, GPT-OSS-120B), with or without an ICL example. We also evaluated the attention-based method on GSM8K. 

\begin{table*}[t]
\centering
\caption{AUPC ($\uparrow$) of top-performing attribution methods on other datasets.}
\label{tab:aupc-all}
\small
\begin{tabular}{lcccc}
\toprule
Algorithm & DS-Llama-8B & DS-Qwen-14B & DS-Qwen3-8B & Qwen3-8B \\
\midrule
\multicolumn{5}{c}{\textit{--- MATH500 ---}} \\
AttriCoT & \underline{\textbf{265.53}} $\pm$ 4.87 & \underline{\textbf{256.07}} $\pm$ 4.74 & \textbf{98.38} $\pm$ 1.87 & \textbf{147.56} $\pm$ 3.45 \\
AttriCoT-2x & 265.50 $\pm$ 4.84 & 255.83 $\pm$ 4.68 & \underline{99.20} $\pm$ 1.87 & \underline{148.40} $\pm$ 3.46 \\
TA-KL (no log) & 234.93 $\pm$ 4.62 & 221.29 $\pm$ 4.76 & 84.16 $\pm$ 2.26 & 134.42 $\pm$ 3.91 \\
Prompted (best same-size) & 181.33 $\pm$ 4.06 & 143.61 $\pm$ 3.02 & 44.86 $\pm$ 1.29 & 74.96 $\pm$ 2.46 \\
Prompted-ICL (GPT-OSS) & 179.94 $\pm$ 3.57 & 170.08 $\pm$ 3.62 & 52.47 $\pm$ 1.38 & 83.09 $\pm$ 2.38 \\
\midrule
\multicolumn{5}{c}{\textit{--- MMLU-Pro ---}} \\
AttriCoT & \textbf{253.51} $\pm$ 2.67 & \textbf{264.42} $\pm$ 2.82 & \textbf{131.87} $\pm$ 1.75 & \textbf{139.61} $\pm$ 1.82 \\
AttriCoT-2x & \underline{254.44} $\pm$ 2.66 & \underline{266.50} $\pm$ 2.82 & \underline{132.77} $\pm$ 1.75 & \underline{140.61} $\pm$ 1.82 \\
TA-KL (no log) & 223.18 $\pm$ 2.51 & 228.23 $\pm$ 2.72 & 118.84 $\pm$ 1.86 & 127.59 $\pm$ 2.02 \\
Prompted (best same-size) & 166.38 $\pm$ 2.28 & 160.66 $\pm$ 2.06 & 61.52 $\pm$ 0.82 & 58.02 $\pm$ 0.97 \\
Prompted-ICL (GPT-OSS) & 150.09 $\pm$ 1.81 & 178.19 $\pm$ 2.32 & 56.00 $\pm$ 0.72 & 67.03 $\pm$ 1.06 \\
\midrule
\multicolumn{5}{c}{\textit{--- GPQA ---}} \\
AttriCoT & \textbf{166.81} $\pm$ 5.61 & \textbf{203.01} $\pm$ 6.65 & \textbf{65.21} $\pm$ 2.41 & \textbf{84.86} $\pm$ 3.22 \\
AttriCoT-2x & \underline{167.37} $\pm$ 5.61 & \underline{204.55} $\pm$ 6.70 & \underline{65.90} $\pm$ 2.41 & \underline{85.53} $\pm$ 3.22 \\
TA-KL (no log) & 149.17 $\pm$ 5.66 & 175.97 $\pm$ 6.75 & 54.26 $\pm$ 2.59 & 70.34 $\pm$ 3.67 \\
Prompted (best same-size) & 119.02 $\pm$ 4.47 & 112.31 $\pm$ 4.09 & 34.52 $\pm$ 1.40 & 38.89 $\pm$ 1.99 \\
Prompted-ICL (GPT-OSS) & 100.81 $\pm$ 3.63 & 119.12 $\pm$ 4.36 & 43.35 $\pm$ 1.36 & 38.38 $\pm$ 1.84 \\
\midrule
\multicolumn{5}{c}{\textit{--- ZebraLogic ---}} \\
AttriCoT & \textbf{277.11} $\pm$ 5.25 & \textbf{347.59} $\pm$ 5.20 & \textbf{174.17} $\pm$ 2.04 & 291.73 $\pm$ 3.06 \\
AttriCoT-2x & \underline{277.66} $\pm$ 5.23 & \underline{348.71} $\pm$ 5.18 & \underline{174.75} $\pm$ 2.02 & 291.94 $\pm$ 2.99 \\
TA-KL (no log) & 273.68 $\pm$ 4.88 & 335.28 $\pm$ 4.76 & 173.69 $\pm$ 1.98 & \underline{\textbf{292.42}} $\pm$ 3.17 \\
Prompted (best same-size) & 211.19 $\pm$ 4.21 & 167.67 $\pm$ 2.51 & 102.90 $\pm$ 1.11 & 156.24 $\pm$ 2.65 \\
Prompted-ICL (GPT-OSS) & 126.37 $\pm$ 2.12 & 157.67 $\pm$ 2.13 & 85.55 $\pm$ 0.73 & 135.95 $\pm$ 1.73 \\
\bottomrule
\end{tabular}
\end{table*}

Table~\ref{tab:aupc_gsm8k} shows the AUPC values achieved by all methods for the four CoT models that we used. Each table entry shows the mean AUPC and standard error of the mean over the dataset samples. 
\begin{itemize}[nosep,left=0pt]
    \item \textbf{AttriCoT} achieves the highest AUPC values among methods that use $S+T$ forward passes or generations. AttriCoT-2x, which uses $2(S+T)$ forward passes, offers a small improvement. 
    \item \textbf{Thought Anchors-KL}: Between the two variants, no log transform is consistently better, so we use it in the remaining experiments. 
    \item \textbf{Prompting}: Among the prompting methods that prompt a same-architecture model (the ``self'' and ``counterpart'' rows in Table~\ref{tab:aupc_gsm8k}), the best-performing variant depends on the CoT model. For DS-Llama-8B and DS-Qwen-14B, providing an ICL example is better, whereas for DS-Qwen3-8B and Qwen3-8B, ICL does not help. The non-distilled counterpart is significantly better for DS-Qwen-14B, while for DS-Qwen3-8B, it is narrowly beaten by the CoT model itself. (At the time of writing, we were not successful in prompting the counterpart of DS-Llama-8B, namely Llama-3.1-8B-Instruct, for this task.) For GPT-OSS-120B, using ICL is consistently better. 
    \item \textbf{Attention-based}: Table~\ref{tab:aupc_gsm8k} shows that attention intervention is mostly competitive with the prompting baselines at much lower inference cost (the attention interventions plus shallow network inference versus full generative calls). However, its performance is not better than the strongest prompting methods and is far below that of the causal intervention methods, AttriCoT and Thought Anchors-KL. For these reasons, we did not evaluate attention intervention on the other datasets and leave further development of a white-box method to future work.
\end{itemize}

Table~\ref{tab:aupc-all} compares the AUPC of the strongest variants in each category from Table~\ref{tab:aupc_gsm8k} on the remaining datasets. ``Prompted (best same-size)'' refers to the best same-architecture model as discussed above. The patterns remain consistent: AttriCoT is almost always the best among $(S+T)$-complexity methods, AttriCoT-2x almost always yields a small improvement, and Thought Anchors-KL is the closest competitor. On ZebraLogic, Thought Anchors-KL is in fact the best performer for the Qwen3-8B model. 

Overall, these results show that AttriCoT is consistently the attribution method that is most behaviorally faithful to the CoT model, across 4 CoT models and 5 datasets covering multiple domains.

\section{Analysis of Attribution Scores}
\label{sec:analysis}

\begin{figure*}[t]
\centering
\begin{subfigure}[b]{0.32\textwidth}
    \centering
    \includegraphics[width=\textwidth]{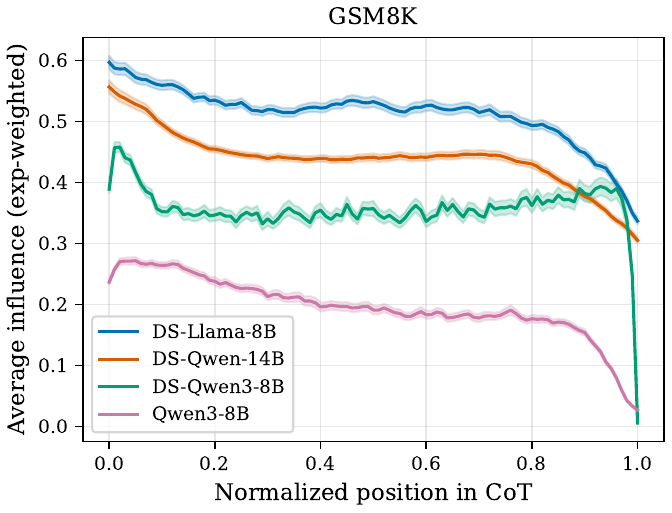}
    \caption{GSM8K}
    \label{fig:influence-gsm8k}
\end{subfigure}
\hfill
\begin{subfigure}[b]{0.32\textwidth}
    \centering
    \includegraphics[width=\textwidth]{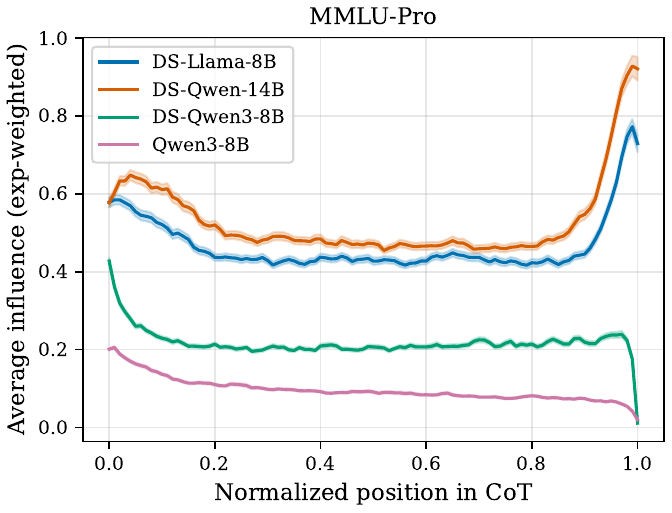}
    \caption{MMLU-Pro}
    \label{fig:influence-mmlu}
\end{subfigure}
\hfill
\begin{subfigure}[b]{0.32\textwidth}
    \centering
    \includegraphics[width=\textwidth]{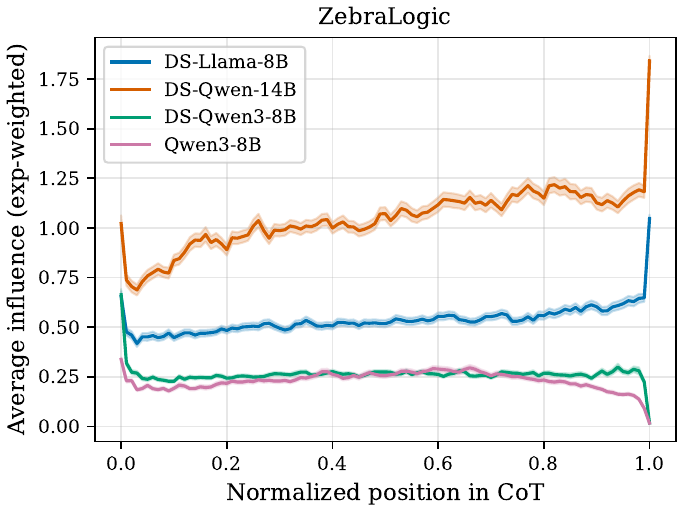}
    \caption{ZebraLogic}
    \label{fig:influence-zebra}
\end{subfigure}
\caption{Exponentially-weighted average influence of CoT units on subsequent units as a function of normalized position in the CoT.}
\label{fig:influence-main}
\end{figure*}

In this section, we perform an analysis of the attribution score matrices ($\mA$ and $\mB$) produced by AttriCoT. This analysis serves both to introduce metrics that summarize the score matrices (Section~\ref{sec:analysis:metrics}) as well as to uncover differences among the LRMs and datasets that we have used (Section~\ref{sec:analysis:results}).

\subsection{Metrics}
\label{sec:analysis:metrics}

We compute the following metrics. Equations for the metrics are provided in Appendix~\ref{sec:appendix-metrics}.

\paragraph{Score density:} The mean of absolute attribution scores over input attribution matrix $\mA$ and the upper-triangular part of output attribution matrix $\mB$.

\paragraph{Average influence:} For each CoT unit, corresponding to a row in $\mB$, the mean absolute score over the part of the row to the right of the main diagonal. This represents the CoT unit's average influence on subsequent output units. We compute both the unweighted average over subsequent units as well as exponentially-weighted averages.

\paragraph{Importance by distance:} For a given distance $d$ in steps between CoT units, corresponding to the $d$th diagonal above the main diagonal in $\mB$, the mean absolute score over the diagonal. This measures the decay in importance of one CoT step to a later step as their separation increases. We compute two versions, one as a function of absolute distance $d$, the other a function of normalized distance $d / (T_{\text{CoT}} - 1)$, where $T_{\text{CoT}}$ is the number of CoT steps (excluding final answer).

\paragraph{Input fraction/ratio:} For each output unit, corresponding to a column in each of $\mA$ and $\mB$, the sum (for input fraction) or average (for input ratio) of absolute scores in the column of $\mA$, divided by the sum (input fraction) or average (input ratio) over the columns of both $\mA$ and $\mB$. This measures the relative importance of input units to the output unit of interest compared to previous output units.

\paragraph{Entropy:} For each output unit, we first concatenate the corresponding columns of $\mA$ and $\mB$ and normalize so that the absolute sum over the concatenated column is $1$. Then we compute the entropy of this normalized column, and normalize the entropy by the entropy of the corresponding uniform distribution. This measures how concentrated are the attributions of the output unit, where significant dependence on only a few prior units would result in low entropy, and diffuse dependence would result in high entropy.
    
To summarize, average influence aggregates over rows of the matrices, importance by distance aggregates over diagonals, input fraction and entropy over columns, and density over the entire matrices.

\subsection{Results}
\label{sec:analysis:results}

We present results in this section for the average influence (exponentially-weighted) and input ratio metrics, on the GSM8K, MMLU-Pro, and ZebraLogic datasets. The full set of results can be found in Appendix~\ref{sec:expt:results_addl:analysis}. 

\paragraph{Average influence.} Figure~\ref{fig:influence-main} plots the exponentially-weighted average influence of CoT units as a function of their normalized position in the CoT. We discuss the preference for using exponential weighting instead of the unweighted average in Appendix~\ref{sec:expt:results_addl:analysis}. The curves represent the mean over samples in the dataset while the shading indicates one standard error above and below the mean. There are interesting differences between the four models and three datasets included in Figure~\ref{fig:influence-main}. First, the curves for DS-Llama-8B and DS-Qwen-14B are higher than for the other two models, which is in line with their differences in terms of the score density metric (Table~\ref{tab:density}). The middle range of each curve is approximately flat, suggesting constant influence, while the distinguishing features occur at the beginnings and ends. For DS-Llama-8B and DS-Qwen-14B, exponentially-weighted average influence decreases at the end on the GSM8K dataset but increases at the end for MMLU-Pro and ZebraLogic. For DS-Qwen3-8B, the curves show a consistent pattern of decreasing at the beginning and again at the end. Qwen3-8B also shows a decrease at the end, but not much of a decrease at the beginning on GSM8K.

\begin{figure*}[t]
\centering
\begin{subfigure}[b]{0.32\textwidth}
    \centering
    \includegraphics[width=\textwidth]{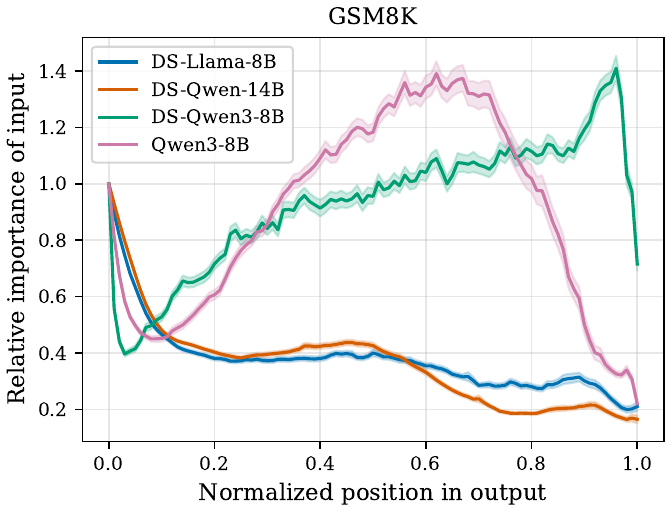}
    \caption{GSM8K}
    \label{fig:input_ratio-gsm8k}
\end{subfigure}
\hfill
\begin{subfigure}[b]{0.32\textwidth}
    \centering
    \includegraphics[width=\textwidth]{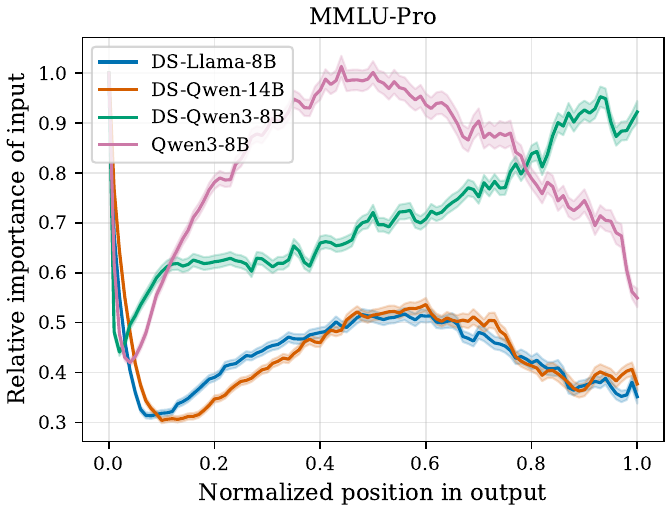}
    \caption{MMLU-Pro}
    \label{fig:input_ratio-mmlu}
\end{subfigure}
\hfill
\begin{subfigure}[b]{0.32\textwidth}
    \centering
    \includegraphics[width=\textwidth]{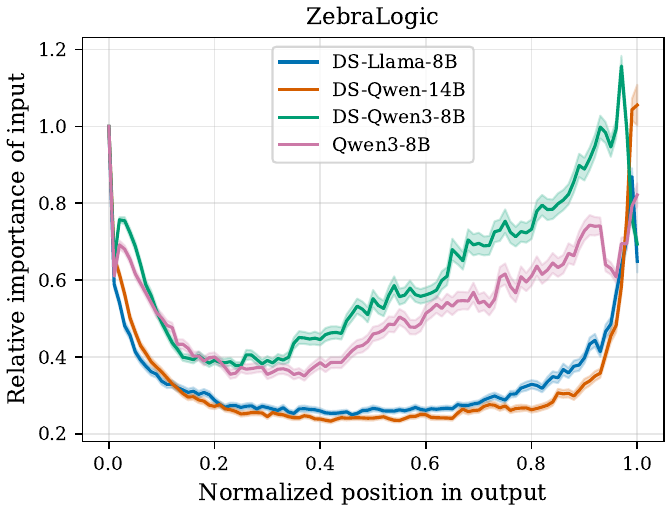}
    \caption{ZebraLogic}
    \label{fig:input-ratio-zebra}
\end{subfigure}
\caption{Relative importance of input units as a function of normalized position in output.}
\label{fig:input-ratio-main}
\end{figure*}

\paragraph{Input ratio.} Figure~\ref{fig:input-ratio-main} plots the input ratio metric for output units as a function of their normalized position (again mean and standard error over dataset samples). 
As mentioned above, input ratio is the ratio between the average attribution score of input units and that of all prior units. 
A high {input ratio} means therefore that input units contribute significantly to the output unit of interest. 
We first note that across the plots, the {input ratio} is high for the first units (around position $x=0$). This is expected if we assume that the first output units rely heavily on the input as the model digests and reformulates the problem. (We plan to corroborate this and subsequent interpretations in future work.) Generally across models and datasets, the input ratio drops gradually. We interpret this as the model gradually relying less on the input as it explores and builds its own reasoning. After this phase (around $x \in [0.2,0.4]$), the input ratio grows again, peaking around the mid to late reasoning stage ($x \in [0.5,0.8]$). We interpret this phenomenon as the model referring back to the input as it reaches an answer, which is a classic behavior of models: once an answer is reached, models start verifying it and putting it in the context of the input. We also note that the peak appears later for more complex problems (i.e., ZebraLogic - Figure~\ref{fig:input-ratio-zebra}, and also MATH500 - Figure~\ref{fig:input-ratio-all-math500}, GPQA - Figure~\ref{fig:input-ratio-all-gpqa} in the appendix). This observation aligns with the behavior of LRMs: more computation is needed to shape answers on complex problems, which delays verification. 
Similar observations as above hold for the input fraction metric (Figure~\ref{fig:input-fraction-all} in Appendix~\ref{sec:expt:results_addl:analysis}), albeit less clearly because of how input fraction is computed.

\section{Conclusion}

Treating a CoT trace as a causal object provides a useful frame for analysis of a model's thought process. Our proposed method, AttriCoT, has pursued this view by decomposing a CoT trace into user-defined units and then formulating and fitting structural equations over the units. The resulting attributions take the form of pairwise importance scores that describe how each unit's log probability depends on prior units. Evaluation on five datasets and four reasoning models shows that AttriCoT produces attributions that are consistently more faithful to model behavior than alternative methods. Metrics derived from AttriCoT's score matrices show suggestive patterns, including the special influence of initial and final CoT steps and models possibly referring back to their inputs in mid-to-late reasoning stages. 

Some limitations of our work include the limited expressiveness of linear structural equations and the simplicity of the default leave-one-out intervention. 
As next steps, we aim to further develop computationally cheap white-box methods and investigate how the structure of reasoning depends on the model family or trace features (e.g., answer correctness, problem difficulty, types of reasoning steps, etc.).

\section*{Acknowledgements}
We thank Eitan Farchi for initial discussions that led to this work and the Mechanistic Interpretability Workshop reviewers for their detailed and helpful comments.

\bibliography{attricot}
\bibliographystyle{icml2026}

\newpage
\appendix
\onecolumn

\section{Related Work} \label{sec:appendix-related-work}

This section highlights the existing methods in the literature, what they aim to do, and how they work. Table \ref{tab:related-works-v2} presents an overview of existing CoT attribution methods \cite{zaman2025chainofthoughtreallyexplainabilitychainofthought}.

\begin{table}[ht]
    \scriptsize
    \centering
    \begin{tabular}{lccccccp{4cm}}
    \toprule
       \textbf{Method} & \textbf{Ref.} & \textbf{Type} & \textbf{Intervention} & \textbf{Causal} & \textbf{Granularity} & \textbf{Access} \\
    \midrule
        Biasing Features & \citep{turpin2023languagemodelsdontsay} & Input perturbation & \checkmark &  & Full CoT & Black-box  \\

        \midrule

        Counterfactual Edits & \citep{atanasova2023faithfulnesstestsnaturallanguage} & CoT perturbation & \checkmark &  & Step/Token & Black-box  \\

        \midrule

        FUR (*) & \citep{tutek-etal-2025-measuring} & Unlearning & \checkmark &  & Step & White-box  \\

        \midrule 

        CRV & \citep{zhao2026verifyingchainofthoughtreasoningcomputational} & Circuit analysis & \checkmark & \checkmark & Step & White-box  \\

        \midrule 

        NLDD & \citep{ye2026mechanisticevidencefaithfulnessdecay} & Counterfactual & \checkmark & \checkmark & Step & White-box   \\

        \midrule

        - & \citep{zaman2025chainofthoughtreallyexplainabilitychainofthought} & Causal intervention & \checkmark & \checkmark & Step & Black-box  \\

        \midrule

        FRODO & \citep{paul2024makingreasoningmattermeasuring} & Causal Intervention & \checkmark & \checkmark & Step & Black-box  \\

        \midrule

        Thought Branches & \citep{macar2025thoughtbranchesinterpretingllm} & Causal Intervention & \checkmark & \checkmark & Step & Black-box  \\

        \midrule

        Thought Anchors & \citep{bogdan2025thoughtanchorsllmreasoning} & Causal Intervention & \checkmark & \checkmark & Step & Both  \\

        \midrule

        CC-SHAP & \citep{parcalabescu2024measuringfaithfulnessselfconsistencynatural} & Attribution &  &  & Token & White-box \\

        \midrule

        - & \citep{minegishi2025topologyreasoningunderstandinglarge} & Pattern analysis &  &  & Step & White-box  \\

        \midrule

        - & \citep{parcalabescu2024measuringfaithfulnessselfconsistencynatural} & Benchmark &  &  & Full CoT & Black-box  \\

        \midrule

        \textbf{AttriCoT} (Ours) & - & Causal intervention & \checkmark & \checkmark & Flexible & Black-box \\
        
    \bottomrule
    \end{tabular}
    \vspace{0.1cm}
    \caption{\small Overview of CoT attribution and faithfulness methods. \textbf{Intervention:} Whether the method perturbs the CoT or the input. \textbf{Causal:} Whether the method suggest a causal framework (e.g. activation patching). \textbf{Granularity:} The degree at which the CoT is treated (Full CoT considers input/output sets, step/tokens intervene inside the CoT). \textbf{Access:} Whether the method requires the model's hidden-states (activations or logprob - White-box), or just the text generated (Black-box).}
    \label{tab:related-works-v2}
\end{table}

\textbf{Faithfulness of LRMs.} Faithfulness has two different meanings in the literature in the context of LRMs. First, \emph{``representational''} faithfulness refers to how much the verbalized chain of thought is representative of the model's true hidden computations \cite{barez2025chain}. Second, faithfulness also refers to how much verbalized steps are contributing to the final answer of the model \cite{zaman2025chainofthoughtreallyexplainabilitychainofthought}. Our work is agnostic of the \emph{``representational''} faithfulness, and operates on a projection of the internal state of the model. It is aligned with the second notion because it exactly addresses the contributions of CoT steps to later steps and the final answer. These dependencies are a meaningful measure of LLM, even when the CoT text does not verbalize internal reasoning \cite{zaman2025chainofthoughtreallyexplainabilitychainofthought}. Rather than measuring how well the generated text aligns with the model internal representation of the reasoning process, AttriCoT identifies which reasoning steps cause subsequent steps in the reasoning trace. In other words, our work aims to reveal if the generated CoT has a coherent causal structure.

\textbf{Perturbation-based, non-causal methods.} The earliest work on CoT attribution are a set of methods that intervenes on the input of the CoT to measure the change in the model's output. First, \citet{turpin2023languagemodelsdontsay} suggest a set of techniques to inject hints in the input and observe how the output is affected. Similarly, \citet{atanasova2023faithfulnesstestsnaturallanguage} insert tokens in the input and manipulates the CoT to then inspect the changes that it implies. Further, CC-SHAP \cite{parcalabescu2024measuringfaithfulnessselfconsistencynatural} applies the Shapley-values to the attribution of the CoT at a token level. However, they measure a correlation with the final output, rather than evaluating causality. 

\textbf{White-box causal methods.} Contrastingly, a set of existing methods evaluates the causality of reasoning traces using the model's internals. CRV \cite{zhao2026verifyingchainofthoughtreasoningcomputational} represent the CoT using hidden-state representation of the model obtained with transcoders. They detect reasoning errors, and uses activation patching to uncover causality of the CoT. NLDD \cite{ye2026mechanisticevidencefaithfulnessdecay} corrupts individual reasoning steps and mesuare how the confidence of the model though its hidden state is affected. 

\textbf{Black-box step-level causal methods.} Recent work suggested black-box methods, evaluating the causality of a CoT at a step-level. These methods perturb individual reasoning steps and evaluate the impact of such edits on the output or on the final answer. FRODO \cite{paul2024makingreasoningmattermeasuring} is one of the first paper introducing the Causal Mediation Analysis (CMA). They compare direct and indirect effects of step perturbations on the model's output. Thoughts Branches \cite{macar2025thoughtbranchesinterpretingllm} also introduce a causal framework to attribute the CoT. The authors re-generate the CoT from different steps in the CoT, and measures the final answer distribution. 

The closest work to ours is Thought Anchors, specifically their sentence-to-sentence masking method \citep[Sec.~5, App.~M]{bogdan2025thoughtanchorsllmreasoning}. This method also attributes every unit in the CoT to prior ones, producing a matrix of attribution scores (their units can only be sentences however). It does so by masking each sentence, either by suppressing all attention to it (white-box) or removing the sentence from the input (black-box), and then performing a forward pass to determine the effects. The difference however is that Thought Anchors quantifies effects on a downstream sentence using KL divergences between token distributions. This is not local in the same way as AttriCoT because it does not measure how the given generated sentence was affected. Instead, it measures the change in token distributions at the position of the sentence. Another difference is that Thought Anchors is limited to leave-one-sentence-out interventions, whereas AttriCoT fits a structural causal model and can thus exploit information from additional interventions if available. \citet{bogdan2025thoughtanchorsllmreasoning} also propose a resampling method, which is more distant from AttriCoT in that it does not yield unit-to-unit attributions, is even less local due to resampling CoTs, and is computationally expensive because of resampling.

\section{Attention Intervention Method}
\label{sec:attention_intervention}

\subsection{Attention intervention and attention output recomputation}
The generation of a token (which we will refer to as the target token) depends on previous tokens through the attention mechanism. The output of each attention head is given by 
\begin{equation}\label{eq:attn_head}
    \mathbf{a} = \sigma\left(\frac{\mathbf{q} \mathbf{K}^T}{\sqrt{d_K}}\right) \mathbf{V} = \sigma(\mathbf{s}) \mathbf{V},
\end{equation}
where $\mathbf{q} \in \mathbb{R}^{1\times d_K}$ is the query vector corresponding to the target token, $\mathbf{K} \in \mathbb{R}^{n\times d_K}$ is the matrix of key vectors for the $n$ previous tokens, $\mathbf{V} \in \mathbb{R}^{n\times d_K}$ is the corresponding matrix of value vectors, $\mathbf{s} \in \mathbb{R}^{1\times n}$ is the vector of $\mathbf{q} \mathbf{k}^T$ inner products (``attention scores''), and $\sigma$ denotes the softmax operation. 

We consider for illustration the case where the target token is the first token of the $t$th output unit $\vy_t$. In this case, we segment the previous token sequence into units $(\vx_1, \dots, \vx_S, \vy_1, \dots, \vy_{t-1})$ as before. The vector $\mathbf{s}$ and matrix $\mathbf{V}$ may be partitioned accordingly so that \eqref{eq:attn_head} can be rewritten as 
\begin{equation}\label{eq:attn_head_partitioned}
    \mathbf{a} = \sigma\left(
    \begin{bmatrix}
        \mathbf{s}_{\vx_1} & \cdots & \mathbf{s}_{\vx_S} & \mathbf{s}_{\vy_1} & \cdots & \mathbf{s}_{\vy_{t-1}}
    \end{bmatrix}\right)
    \begin{bmatrix}
        \mathbf{V}_{\vx_1} \\ \vdots \\ \mathbf{V}_{\vy_{t-1}}
    \end{bmatrix}
    = \sum_{s=1}^S \mathbf{p}_{\vx_s} \mathbf{V}_{\vx_s} + \sum_{\tau=1}^{t-1} \mathbf{p}_{\vy_\tau} \mathbf{V}_{\vy_\tau},
\end{equation}
where we have defined \emph{attention weights} $\mathbf{p} = \sigma(\mathbf{s})$.

We now approximate the effect of removing a unit, say the first CoT unit $\vy_1$, by removing the corresponding subvector $\mathbf{s}_{\vy_1}$ and submatrix $\mathbf{V}_{\vy_1}$ from \eqref{eq:attn_head_partitioned}. This corresponds to setting the corresponding attention weights $\mathbf{p}_{\vy_1}$ to zero. The effect on the remaining softmax outputs $\mathbf{p}_{\vx_1}, \dots, \mathbf{p}_{\vx_S}, \mathbf{p}_{\vy_2}, \dots, \mathbf{p}_{\vy_{t-1}}$ is just a renormalization, so that the modified attention output becomes 
\begin{equation}\label{eq:attn_head_perturbed}
    \mathbf{a}' = \sum_{s=1}^S \frac{\mathbf{p}_{\vx_s}}{1 - \mathbf{1}^T \mathbf{p}_{\vy_1}} \mathbf{V}_{\vx_s} + \sum_{\tau=2}^{t-1} \frac{\mathbf{p}_{\vy_\tau}}{1 - \mathbf{1}^T \mathbf{p}_{\vy_1}} \mathbf{V}_{\vy_\tau},
\end{equation}
where $\mathbf{1}$ is a vector of ones. (This is an approximation because the removal of $\vy_1$ also modifies the key and value vectors of subsequent thoughts $\vy_2, \dots, \vy_{t-1}$, but we neglect these second-order effects.) 
The above intervention is repeated on all attention heads within a Transformer block of the model, resulting in a perturbed output like \eqref{eq:attn_head_perturbed} from each attention head.

\subsection{Shallow network for propagation to logits}

Our proposal is to use perturbed attention outputs as in \eqref{eq:attn_head_perturbed} as the starting point to estimate the downstream effect on the log probabilities of the tokens of a target unit $\vy_t$. These log probabilities are determined by the corresponding logits, so the task is to propagate perturbed attention head outputs \eqref{eq:attn_head_perturbed} to perturbed logits. We consider doing so using a ``shallow network.'' 

We first discuss an ``untrained'' version in which the shallow network consists of modules in the LLM along the path that connects attention outputs to logits, with the same (frozen) parameter values. In the case where the attention intervention is applied to the model's last Transformer block, then the shallow network is the complete set of modules between attention outputs and logits, namely: the output projection (``o-projection'') that combines attention head outputs; addition of the o-projected attention output to the residual stream; the subsequent MLP layer, with residual connection and layer normalization; the LLM's language modeling head, also preceded by a layer normalization. If the Transformer block is not the last one, then we still take the o-projection and MLP layer from the same Transformer block, but then skip over all subsequent Transformer blocks and just attach the language modeling head, similar to the LogitLens technique \cite{nostalgebraist2020logitlens} in mechanistic interpretability. The choice of Transformer block where the intervention is applied is a hyperparameter of the method and we experimented with LLM-dependent ranges of block indices (please see Table~\ref{tab:white-box-blocks} for the results).

We also tested a ``trained'' version, in which the shallow network is initialized from the untrained version. We then train the shallow network using AttriCoT's computed log probabilities as supervision. Specifically, for a given sequence of units and a perturbation to some of the units, we can compute the log probabilities of all tokens as AttriCoT does by performing a full forward pass. We also have the shallow network's prediction of these log probabilities, obtained by performing the corresponding intervention on attention weights and propagating through the shallow network. We compare the two sets of log probabilities using mean squared error as the loss function. 

We trained shallow networks using a deliberately small set of 10 samples (8 for training, 2 for validation) from the GSM8K dataset. This number was chosen because it represents less than 1\% of the full test set, with the idea that the cost of training could be amortized over evaluation on the rest of the dataset. Moreover, each sample actually yields $S + T + 1$ unit sequences, one perturbed sequence for each of its $S + T$ units plus an unperturbed sequence, and each of these sequences can be a long sequence of tokens. Hence the amount of training data is larger than it might appear.

\section{Experimental Design} \label{sec:appendix-experiment-design}

\subsection{Description of selected datasets and models}

\begin{table}[ht]
    \small
    \centering
    \begin{tabular}{cccccc}
    \toprule
    \multirow{2}{*}{\textbf{Domain}} & \multirow{2}{*}{\textbf{Ref.}} & \multirow{2}{*}{\textbf{Name}} & \multirow{2}{*}{\textbf{Size}} & \multicolumn{2}{c}{\textbf{Type of evaluation}} \\
    \cmidrule(lr){5-6} 
     &  &  &  & Math-Verify\footnote{\url{https://github.com/huggingface/Math-Verify}} & MCQ Prompting \\
    \midrule
       \multirow{2}{*}{Maths} & \cite{cobbe2021gsm8k} & GSM8K & $1319$ & \cmark &  \\
        & \cite{hendrycks2021measuringmathematicalproblemsolving} & MATH500 & $500$ & \cmark &  \\
        \midrule
        \multirow{2}{*}{Multi-domain} & \cite{rein2023gpqagraduatelevelgoogleproofqa} & GPQA-Diamond & $198$ &  & \cmark \\
        & \cite{wang2024mmluprorobustchallengingmultitask} & MMLU-Pro & $1400$ &  & \cmark \\
        \midrule
        \multirow{1}{*}{Logic} & \cite{zebralogic2025} & ZebraLogic & $564$ &  & \cmark \\
    \bottomrule
    \end{tabular}
    \caption{Overview of selected reasoning datasets}
    \label{tab:reasoning-datasets}
\end{table}

\paragraph{Selection of datasets.} To evaluate AttriCoT and the baselines, we selected 5 reasoning benchmarks across 3 different domains. Table \ref{tab:reasoning-datasets} presents the datasets. For all the datasets, we selected the test split that is available. Due to the computational cost of perturbation curve evaluation (see Appendix~\ref{sec:expt:perturb_curve_addl}), we selected 1,400 samples from the full MMLU-Pro dataset by selecting 100 samples from each of the 14 categories in MMLU-Pro. These categories correspond to domains such as History, Law, Computer Science, Maths, or Physics. 

\textbf{ZebraLogic.} The ZebraLogic dataset is complex by nature, and requires significant reasoning efforts compared to other datasets (i.e. models tend to generate more tokens). At the same time, the dataset is significantly large (around $3.2$k samples). Due to computational limitations, we first inferred around 20\% of the first samples of the dataset (i.e. 620 samples). On these samples, we observed that a significant portion of them were not completed. By not completed, we mean that some reasoning traces did not include \texttt{</think>} under their maximum number of tokens. For each sample, we selected it only if all of our selected models completed their reasoning. Pruning the other samples, we obtained $564$ samples. 

\begin{figure}[ht]
    \centering
    \begin{subfigure}{0.95\textwidth}
        \includegraphics[width=\linewidth]{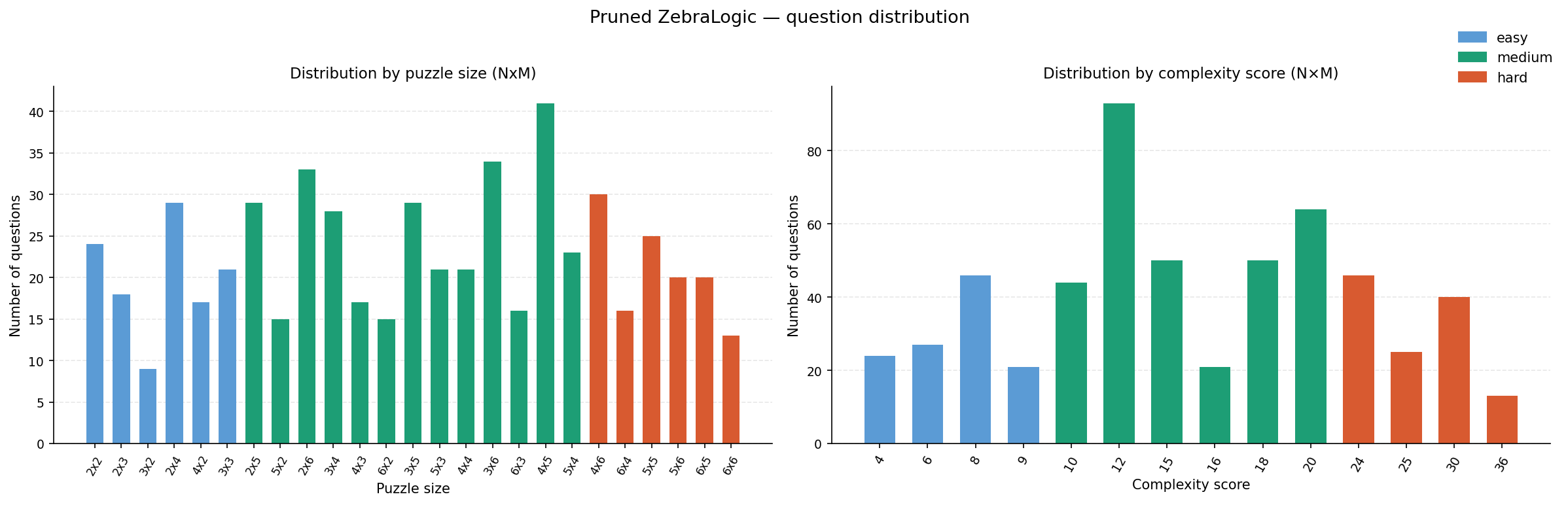}
        \caption{Distribution of puzzle's size}
        \label{fig:distribution-puzzle-size}
    \end{subfigure}
    \hfill
    \begin{subfigure}{0.95\textwidth}
        \includegraphics[width=\linewidth]{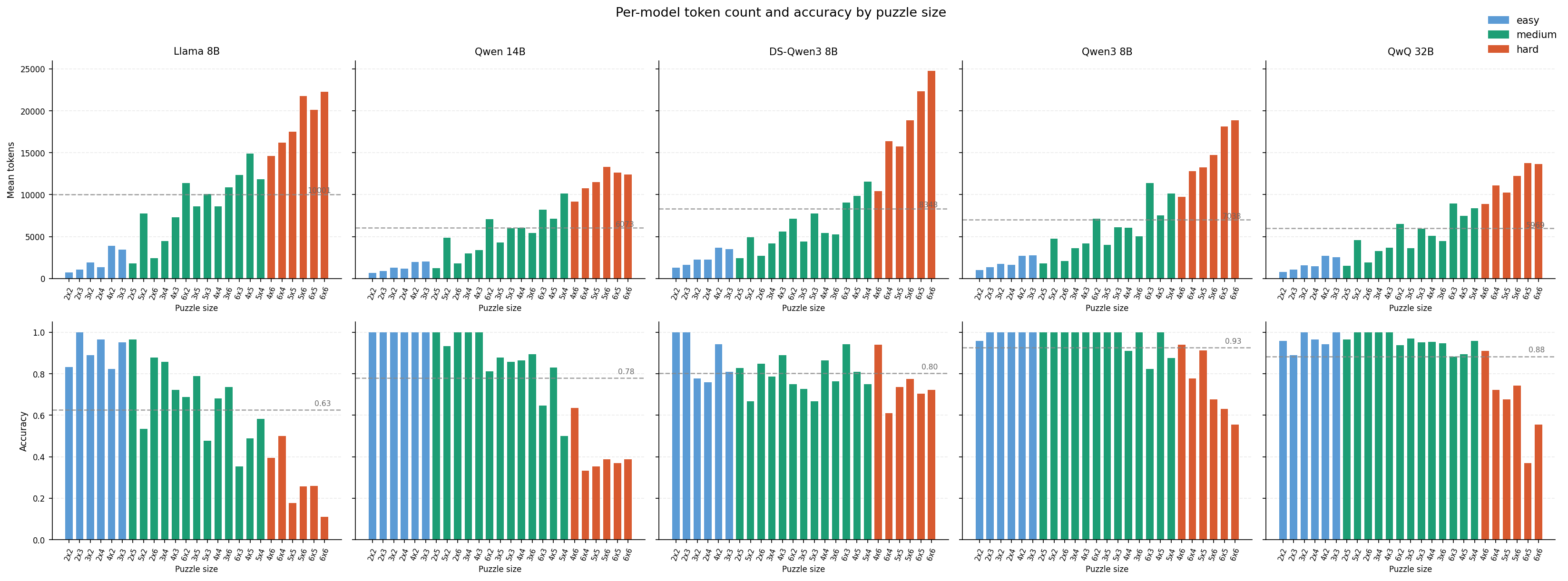}
        \caption{Model's performance per puzzle's size}
        \label{fig:performance-per-complexity}
    \end{subfigure}
    \caption{Analysis of ZebraLogic subset}
    \label{fig:analysis-zebra-logic}
\end{figure}

Next, we analyzed the complexity of the $564$ samples obtained on the ZebraLogic dataset. Such analysis would allow us to ensure a fair distribution of the complexity of the questions composing our subset. Figures \ref{fig:distribution-puzzle-size} and \ref{fig:performance-per-complexity} present the distribution of the puzzle sizes and the performance of the model with respect to the puzzle dimensions. We define the puzzle size by the product of the number of columns $M$ and rows $N$ of each puzzles ($S = M \times N$). Figure \ref{fig:distribution-puzzle-size} shows that the puzzle size is widely spread across our selected samples, namely from $4$ to $36$. 

Further, Figure \ref{fig:performance-per-complexity} shows the performance of models with regards to the puzzle sizes. We observe that the puzzle dimension correlates well with the complexity of the problem. Indeed, the larger the puzzle is, the more token the models generates, and the lower their performance is. To this extend, we arbitrary defined three levels of complexity, namely easy ($S < 10$), medium ($S \in [10,20]$), and hard ($S > 20$).

\textbf{Selection of models.} Table \ref{tab:selected-models} presents the reasoning models that we selected.

\begin{table}[ht]
    \centering
    \scriptsize
    \begin{tabular}{ccccc}
        \toprule
        \textbf{Name} & \textbf{Ref.} & \textbf{Non-reasoning counterpart} & \textbf{Reasoning} \\
        \midrule
        \texttt{Qwen-3-8B} & \cite{qwen3technicalreport} & \texttt{Qwen/Qwen3-8B} & \texttt{Qwen/Qwen3-8B} \\
        \midrule
        \texttt{DS-Qwen3-8B} & \cite{qwen3technicalreport, deepseekai2025deepseekr1incentivizingreasoningcapability} & \texttt{Qwen/Qwen3-8B} & \texttt{deepseek-ai/DeepSeek-R1-0528-Qwen3-8B} \\
        \midrule
        \texttt{DS-Llama-8B} & \cite{grattafiori2024llama3herdmodels,deepseekai2025deepseekr1incentivizingreasoningcapability} & \texttt{meta-llama/Llama-3.1-8B-Instruct} & \texttt{deepseek-ai/DeepSeek-R1-Distill-Llama-8B} \\
        \midrule
        \texttt{DS-Qwen-14B} & \cite{deepseekai2025deepseekr1incentivizingreasoningcapability} & \texttt{Qwen/Qwen-2.5-14B} & \texttt{deepseek-ai/DeepSeek-R1-Distill-Qwen-14B} \\
        \bottomrule
    \end{tabular}
    \caption{Overview of selected reasoning models and their non-reasoning counterparts.}
    \label{tab:selected-models}
\end{table}

\subsection{Generation and segmentation of CoTs}

For each sample of each dataset shown in Table~\ref{tab:reasoning-datasets} (e.g.~1400 samples from MMLU-Pro), we take the prompt as input $\vx$ and generate an output $\vy$ from each of the models in Table~\ref{tab:selected-models}. The prompt may be just a question in the case of the math datasets or also include answer choices in the MCQ datasets. Each output consists of a CoT reasoning trace, terminated by an end-of-thinking \texttt{</think>} token, followed by a ``final answer,'' which may also be of significant length.

\paragraph{Segmentation of the prompt.} For the math datasets (GSM8K and MATH500), the prompt/question $\vx$ was split into sentences $(\vx_1, \dots, \vx_S)$. For the MCQ datasets (MMLU-Pro, GPQA, ZebraLogic), the prompt was split on the newline character \texttt{"\textbackslash n"} into $(\vx_1, \dots, \vx_S)$. In both cases, each segment $\vx_s$ may provide a different piece of information or instruct the model on how to express its answer. For the MCQ datasets, the segments also separate the answer choices (``A. ...'', ``B. ...'', etc.). Segmenting prompts in this way allows differential attribution to elements within the prompt.

\paragraph{Segmentation of the CoT.} 
We segment the CoT reasoning traces by splitting on a delimiter. The resulting segments form the first part of the output unit sequence $(\vy_1, \dots, \vy_T)$. 
In the literature, multiple delimiters have been proposed \citep{zhang2025reasoningmodelsknowtheyre,yao2023treethoughtsdeliberateproblem, lee2025evaluatingstepbystepreasoningtraces, cao2025stepguidedreasoningimproving}. In this paper, we segmented the reasoning traces based on the double-newline delimiter 
\texttt{"\textbackslash n\textbackslash n"} \citep{zhang2025reasoningmodelsknowtheyre}, since reasoning models appear to natively generate these tokens. 

After segmentation, we limit the number of CoT steps to $100$, again because of the computational constraints of perturbation curve computation (Appendix~\ref{sec:expt:perturb_curve_addl}). We do so by merging consecutive steps together. If segmentation produces $T_{\text{CoT}} > 100$ steps, then each step in the merged sequence consists of either $\lfloor T_{\text{CoT}} / 100 \rfloor$ or $\lceil T_{\text{CoT}} / 100 \rceil$ old steps, chosen so that the sum over merged steps equals $T_{\text{CoT}}$. For example if $T_{\text{CoT}} = 140$, 60 of the merged steps consist of 1 old step while 40 consist of 2. When there is a choice between $\lfloor T_{\text{CoT}} / 100 \rfloor$ or $\lceil T_{\text{CoT}} / 100 \rceil$, the one that keeps the cumulative length in characters closer to that of a uniform distribution is chosen. 

\paragraph{Segmentation of the final answer.} We also segment the final answer by splitting on \texttt{"\textbackslash n\textbackslash n"}, since models often produce a summary of their reasoning in their final answer and continue to use the \texttt{"\textbackslash n\textbackslash n"} separator in doing so. In the rare cases where it is needed, we also merge final answer steps into 100 total steps as described above. The sequences of CoT steps and final answer steps together make up the output units $(\vy_1, \dots, \vy_T)$.

\subsection{Additional details on attribution methods}
\label{sec:expt:methods_addl}

\paragraph{AttriCoT}
For the experimental results reported in this paper, we did not use regularization, i.e., $\lambda = 0$. 

For $N > S + T$ interventions (for example $N = 2(S + T)$ for AttriCoT-2x), we followed the approach of the C-LIME algorithm in \citet{monteiro-paes-etal-2025-mexgen} to select interventions. That is, we limit the number of units perturbed at one time to a small integer $K$, and we sample an equal number of perturbations of each cardinality from $1$ to $K$. For AttriCoT-2x, $K = 2$, and hence $S + T$ perturbations are leave-one-out while the other $S + T$ are leave-two-out, sampled uniformly from all $\binom{S+T}{2}$ pairs of units.

\paragraph{Thought Anchors-KL} 
We re-implemented Thought Anchors-KL by starting from the authors' code and making several modifications and improvements:
\begin{enumerate}[nosep,left=5pt]
    \item \textbf{More general units:} We generalized Thought Anchors-KL to operate on the same more general units used by AttriCoT (prompt, CoT, and final answer units) rather than the original's restriction to CoT sentences only. This enables a fairer comparison using the same unit segmentation for both AttriCoT and Thought Anchors. 
    \item \textbf{Batch forward passes:} Like AttriCoT-LOO, Thought Anchors-KL performs leave-one-out perturbations on the unit sequence and then runs forward passes through the LLM. In the case of Thought Anchors-KL, each forward pass computes token distributions at all token positions, rather than just the log probability of the originally generated token in AttriCoT's case. We replaced Thought Anchors' per-perturbation calls to the LLM with a single batch of forward passes, including the original sequence and all perturbed sequences. This batching is more efficient and does not change the result. 
    \item \textbf{Token position matching:} Thought Anchors-KL quantifies the causal effect of a perturbation on a target unit by computing KL divergences between token distributions at every token position in the target unit, and then taking the mean of these KL divergences over the unit. This requires matching token positions in the original sequence with their corresponding token positions in a perturbed sequence. We improved the matching procedure and made it more robust. Thought Anchors' original approach searched for an ``anchor'' of consecutive matching tokens following the unit that was removed in the perturbed sequence. This often failed in our experience because of re-tokenization near the removal boundaries (e.g., newline characters merging with adjacent tokens). Our approach first finds the token ranges corresponding to the target unit in both the original and perturbed sequences. It then runs a greedy two-pointer alignment algorithm on the two token ranges.
    \item \textbf{Log transform:} The Thought Anchors paper \citep{bogdan2025thoughtanchorsllmreasoning} proposed first applying a log transform to KL divergences before taking the mean over a target unit. The authors' code however does not apply a log transform. We evaluated both versions on the GSM8K dataset, as reported in Table~\ref{tab:aupc_gsm8k}, to see which is better for our purpose.
\end{enumerate}

\paragraph{Prompting}
To prompt an LLM to perform CoT attribution, we used the conversation format shown in Figure~\ref{fig:prompting_attribution}. As mentioned in Section~\ref{sec:expt:setup}, we prompt the LLM to attribute output units one at a time, and the example in Figure~\ref{fig:prompting_attribution} is for an output unit $J$. The first user turn is the original prompt (e.g.~a math question), divided into input units $\vx_1, \dots, \vx_S$ (we use $0$-based indexing in the code). The first assistant turn is the LRM's response to the prompt, divided into units $\vy_1, \vy_2, \dots$. Before each unit $\vx_1, \dots, \vx_S, \vy_1, \dots$, we insert a tag of the form \texttt{<uN>}, where the unit index increases in a single sequence across input and output units. The response has also been truncated to the target unit $J$, which we found led to better performance by removing the distraction of later units that cannot affect unit $J$.

\begin{figure}[ht]
\centering
\footnotesize
\begin{adjustbox}{max width=\textwidth}
\begin{tcolorbox}[colback=gray!5, colframe=black, title=Prompting baseline, fonttitle=\bfseries]
\textit{[in-context example if used, following the same format below]}\\
\\
< | User | > <u$0$> \textit{[unit $0$]} <u$1$> \textit{[unit $1$]} ... <u$(S-1)$> \textit{[unit $S-1$]} \\
< | Assistant | > <u$S$> \textit{[unit $S$]} <u$(S+1)$> \textit{[unit $S+1$]} ... <u$(S+J)$> \textit{[unit $S+J$]} \\
< | User | > The above is a conversation that has been divided into units. \\
Each unit is tagged with an identifier in the format <u0>, <u1>, etc. \\ 
For unit <u$(S+J)$>, please list the preceding units that were most important for generating it. \\
List the unit numbers in order from most important to least important, for example [3, 1, 4]. \\
Answer with only the list and do not explain. \\
< | Assistant | > \textit{[response giving attributions as a ranked list, e.g. [1, 5, 9, 2]]}
\end{tcolorbox}
\end{adjustbox}
\caption{Conversation format used by the prompting baseline to attribute output unit $J$.}
\label{fig:prompting_attribution}
\end{figure}

The second user turn is the instruction to the LLM, describing the attribution task and answer format. The LLM then responds with attributions as a ranked list of unit indices. Note that the LLM being prompted to perform attribution may be different from the LRM that generated the response in the first assistant turn, as discussed in Section~\ref{sec:expt:setup}. In particular, we observed that when a DeepSeek-distilled model is prompted to attribute its own response (self-attribution), it first engages in lengthy reasoning before finally outputting a ranked list. Because of this, we allowed the prompted model to generate up to $1000$ new tokens, even though the ranked list is expected to take far fewer tokens. The non-DeepSeek-distilled counterpart models did not have the same issue.

An in-context learning (ICL) example can be prepended to the conversation shown in Figure~\ref{fig:prompting_attribution}, following the same conversation format. We used AttriCoT's attribution results on the same (dataset, CoT model) pair as the source of ICL examples, specifically a sample with the largest number of output units. This choice of data sample allowed us to match the ICL example to the output unit being attributed, i.e., for attributing output unit $J$ as in Figure~\ref{fig:prompting_attribution}, we supply an example of AttriCoT also attributing the $J$th output unit. Since AttriCoT produces numerical scores while we only prompt LLMs for a ranked list, we converted AttriCoT's scores to a ranked list of units with the top $k$ scores. The number $k$ was set to 20\% of the number of units preceding the output unit (e.g.~20\% of $S+J$ in Figure~\ref{fig:prompting_attribution}), which matches the largest $k$ used in perturbation curve evaluation (see the next section). 

\subsection{Perturbation curve evaluation}
\label{sec:expt:perturb_curve_addl}

As mentioned in Section~\ref{sec:expt:setup}, for each output unit $\vy_t$, we compute a perturbation curve that measures the decrease in log probability of $\vy_t$ as a function of the number $k$ of top-ranked prior units that are removed. This decrease in log probability is given by 
\begin{equation}\label{eq:perturb_curve}
    D_t(k) = \frac{1}{\abs{\vy_t}} \log P\left(\vy_t \mid  \vx, \vy_{<t} \right) - \frac{1}{\abs{\vy_t}} \log P\left(\vy_t \mid  \vx^{(k)}, \vy_{<t}^{(k)} \right), 
\end{equation}
where $(\vx^{(k)}, \vy_{<t}^{(k)})$ is the sequence of prior units with the top $k$ removed, and we take the mean of token log probabilities as in \eqref{eq:unit_log_probs} rather than the sum in \eqref{eq:sem_logprob}. Note that the decrease $D_t(k)$ could be negative.

Since we sweep $k$ from $0$ to 20\% of the total number of prior units, $S + t - 1$, each output unit's perturbation curve has a different number of points. 
To average perturbation curves corresponding to different output units, we first convert $k$ to a fraction $k / (S + t - 1)$, and then linearly interpolate each perturbation curve onto a common grid of perturbation fractions (spanning 0\% to 20\%). These interpolated perturbation curves can then be averaged to produce one perturbation curve per data sample, and thus one AUPC value per sample as well. 

It is worth noting that evaluating perturbation curves is more computationally expensive than running any of the attribution algorithms. Since a perturbation curve is computed for each output unit $t = 1,\dots,T$, and the number of perturbations/forward passes is proportional to the number of prior units $S + t - 1$, the total number of forward passes is quadratic: $O(T(S + T))$. It is primarily for this reason that we limited the number of output units to $T = 100$. We believe that this cost is justified in order to thoroughly evaluate CoT attributions, and we emphasize that all of the attribution algorithms have linear cost in $S + T$.

\subsection{Metrics for attribution score matrices}
\label{sec:appendix-metrics}

Here we provide equations for the metrics defined in Section~\ref{sec:analysis:metrics} for summarizing attribution score matrices $\mA \in \mathbb{R}^{S \times T}$ (input $\to$ output) and $\mB \in \mathbb{R}^{T \times T}$ (output $\to$ output, strictly upper triangular). Let $T_{\text{CoT}}$ denote the number of CoT output units (excluding final answer units). We also define the concatenated matrix $\mC = [\mA; \mB] \in \mathbb{R}^{(S+T) \times T}$, which stacks $\mA$ above $\mB$.

\paragraph{Score density.} The mean absolute attribution score over all structurally non-zero entries:
\begin{equation}\label{eq:density}
    \text{density} = \frac{\displaystyle\sum_{s=1}^{S} \sum_{t=1}^{T} \abs{\alpha_{s,t}} \;+\; \sum_{\tau=1}^{T} \sum_{t=\tau+1}^{T} \abs{\beta_{\tau,t}}}{S \cdot T + T(T-1)/2}\,.
\end{equation}

\paragraph{Average influence.} For each CoT unit $\tau = 1, \dots, T_{\text{CoT}}$, the mean absolute attribution from all downstream output units:
\begin{equation}\label{eq:influence}
    I(\tau) = \frac{1}{T - \tau} \sum_{t=\tau+1}^{T} \abs{\beta_{\tau,t}}\,.
\end{equation}
We also define an exponentially-weighted variant with decay rate $r \in [0,1)$:
\begin{equation}\label{eq:influence-exp}
    I_r(\tau) = \frac{1 - r}{1 - r^{T-\tau}}\sum_{d=1}^{T-\tau} r^{d-1}\, \abs{\beta_{\tau,\tau+d}}\,.
\end{equation}
The motivation for the exponentially-weighted variant and the choice of $r$ are discussed in Appendix~\ref{sec:expt:results_addl:analysis}. Both \eqref{eq:influence} and \eqref{eq:influence-exp} are plotted as a function of normalized CoT position $\tau / T_{\text{CoT}}$. 

\paragraph{Importance by distance.} For a distance $d$ in CoT steps, the mean absolute score along the $d$-th super-diagonal of $\mB$:
\begin{equation}\label{eq:locality}
    L(d) = \frac{1}{T_{\text{CoT}} - d} \sum_{\tau=1}^{T_{\text{CoT}}-d} \abs{\beta_{\tau,\tau+d}}\,.
\end{equation}
We plot this both as a function of absolute distance $d$ and of normalized distance $d / (T_{\text{CoT}} - 1)$.

\paragraph{Input fraction.} For each output unit $t$, the fraction of total incoming absolute attribution that originates from input units:
\begin{equation}\label{eq:input-fraction}
    F(t) = \frac{\displaystyle\sum_{s=1}^{S} \abs{\alpha_{s,t}}}{\displaystyle\sum_{s=1}^{S} \abs{\alpha_{s,t}} + \sum_{\tau=1}^{t-1} \abs{\beta_{\tau,t}}}\,.
\end{equation}

\paragraph{Input ratio.} For each output unit $t$, the ratio of mean absolute input attribution to mean absolute attribution over all prior units. Let $c_{j,t}$ denote entry $(j, t)$ of $\mC$; the units prior to output unit $t$ correspond to rows $j = 1, \dots, S + t - 1$ of $\mC$:
\begin{equation}\label{eq:input-ratio}
    R(t) = \frac{\displaystyle\frac{1}{S} \sum_{s=1}^{S} \abs{\alpha_{s,t}}}{\displaystyle\frac{1}{S+t-1} \sum_{j=1}^{S+t-1} \abs{c_{j,t}}}\,.
\end{equation}

\paragraph{Normalized entropy.} For each output unit $t$, we form a distribution from the absolute scores of its predecessors,
\begin{equation}\label{eq:entropy-dist}
    p_j = \frac{\abs{c_{j,t}}}{\displaystyle\sum_{j'=1}^{S+t-1} \abs{c_{j',t}}}, \quad j = 1, \dots, S+t-1,
\end{equation}
and compute the normalized Shannon entropy:
\begin{equation}\label{eq:entropy}
    E(t) = \frac{-\displaystyle\sum_{j=1}^{S+t-1} p_j \ln p_j}{\ln(S+t-1)}\,.
\end{equation}
This metric lies in $[0, 1]$, where low values indicate concentrated attributions and high values indicate diffuse attributions.

\section{Additional Experimental Results}
\label{sec:expt:results_addl}

\subsection{Attention intervention method}
\label{sec:expt:results_addl:white-box}

Table \ref{tab:white-box-blocks} presents the full AUPC performance results on the GSM8K dataset for the attention intervention method of Section~\ref{sec:method:white-box}, including both untrained and trained versions and interventions on different Transformer blocks within each model. Five Transformer block indices were selected to span depths $50\%$--$100\%$ of each model (index 0 is the first block). The block for which the highest AUPC is obtained varies across models: For DS-Llama-8B, it is the middle block (16 out of 32); for DS-Qwen3-8B and Qwen3-8B, it is the block at three-quarters depth; for DS-Qwen-14B, both three-quarters and maximum depth are good. The bold values in Table~\ref{tab:white-box-blocks} are those that are also included in Table~\ref{tab:aupc_gsm8k}. We also observe that while training increases AUPC in 12 of the 20 cases, it improves the best AUPC only for DS-Qwen-14B.

\begin{table}[t]
\centering
\caption{AUPC ($\uparrow$) of the white-box attention intervention method on GSM8K, intervening on different Transformer blocks.}
\label{tab:white-box-blocks}
\begin{tabular}{llcc}
\toprule
Model & Block & Untrained & Trained \\
\midrule
DS-Llama-8B & 15 & \textbf{107.06} $\pm$ 1.25 & \textbf{101.43} $\pm$ 1.35 \\
 & 19 & 106.10 $\pm$ 1.29 & 90.81 $\pm$ 1.29 \\
 & 23 & 89.48 $\pm$ 1.08 & 96.64 $\pm$ 1.12 \\
 & 27 & 83.53 $\pm$ 0.97 & 87.50 $\pm$ 1.02 \\
 & 31 & 90.39 $\pm$ 1.12 & 81.89 $\pm$ 1.05 \\
\midrule
DS-Qwen-14B & 23 & 65.65 $\pm$ 0.96 & 75.21 $\pm$ 1.09 \\
 & 29 & 49.08 $\pm$ 0.78 & 71.74 $\pm$ 0.96 \\
 & 35 & 85.15 $\pm$ 1.13 & \textbf{111.39} $\pm$ 1.19 \\
 & 41 & 68.44 $\pm$ 0.81 & 73.52 $\pm$ 0.86 \\
 & 47 & \textbf{87.36} $\pm$ 1.07 & 86.37 $\pm$ 1.09 \\
\midrule
DS-Qwen3-8B & 17 & 42.72 $\pm$ 0.48 & 39.48 $\pm$ 0.48 \\
 & 21 & 33.91 $\pm$ 0.42 & 50.88 $\pm$ 0.55 \\
 & 26 & \textbf{65.72} $\pm$ 0.61 & \textbf{61.88} $\pm$ 0.58 \\
 & 31 & 48.09 $\pm$ 0.54 & 52.04 $\pm$ 0.55 \\
 & 35 & 31.51 $\pm$ 0.46 & 36.99 $\pm$ 0.54 \\
\midrule
Qwen3-8B & 17 & 54.48 $\pm$ 0.72 & 40.12 $\pm$ 0.66 \\
 & 21 & 42.54 $\pm$ 0.61 & 54.05 $\pm$ 0.73 \\
 & 26 & \textbf{58.31} $\pm$ 0.73 & \textbf{56.02} $\pm$ 0.71 \\
 & 31 & 42.14 $\pm$ 0.59 & 55.63 $\pm$ 0.65 \\
 & 35 & 39.40 $\pm$ 0.59 & 39.82 $\pm$ 0.60 \\
\bottomrule
\end{tabular}
\end{table}

\clearpage

\subsection{Perturbation curves}
\label{sec:expt:results_addl:perturbation}

\begin{figure}[t]
\centering
\begin{subfigure}[b]{0.48\textwidth}
    \centering
    \includegraphics[width=\textwidth]{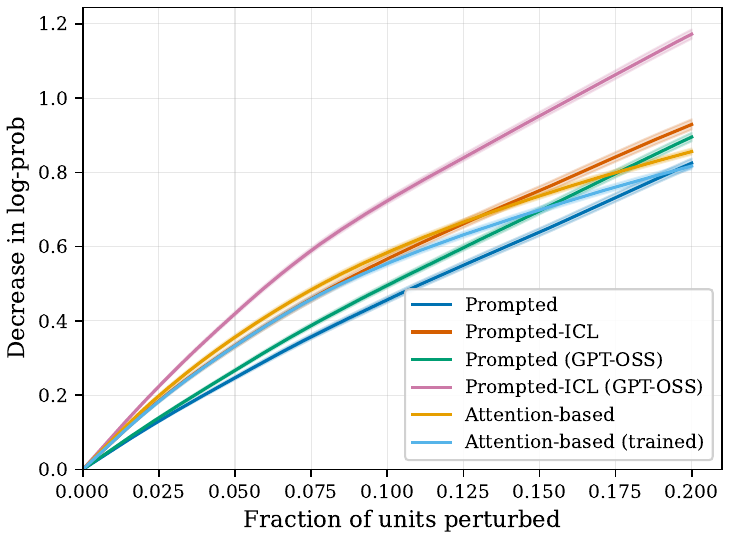}
    \caption{DS-Llama-8B}
    \label{fig:perturb-curves-gsm8k-prompting-ds-llama-8b}
\end{subfigure}
\hfill
\begin{subfigure}[b]{0.48\textwidth}
    \centering
    \includegraphics[width=\textwidth]{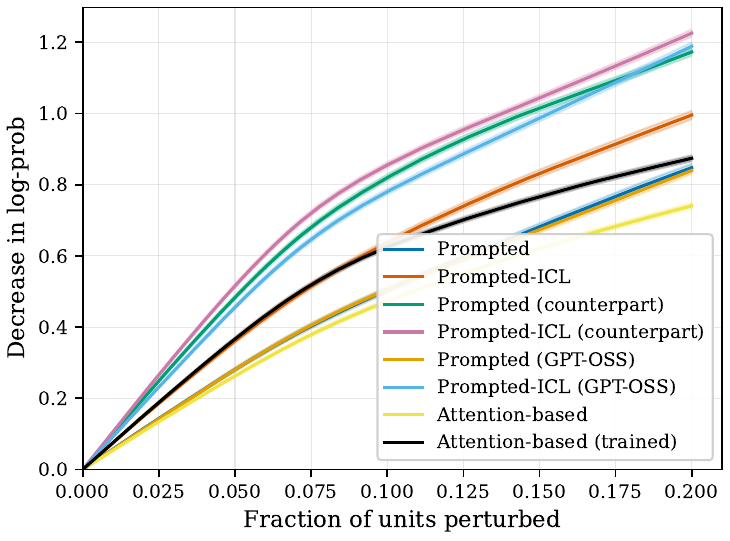}
    \caption{DS-Qwen-14B}
    \label{fig:perturb-curves-gsm8k-prompting-ds-qwen-14b}
\end{subfigure}

\vspace{0.5em}

\begin{subfigure}[b]{0.48\textwidth}
    \centering
    \includegraphics[width=\textwidth]{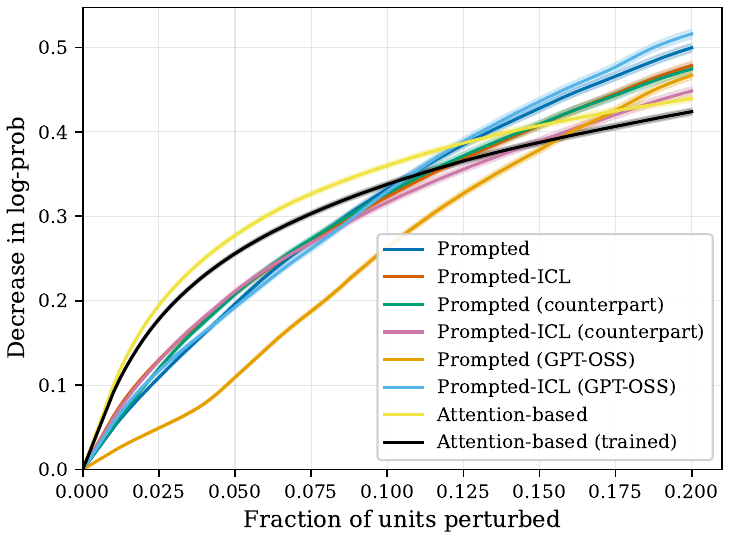}
    \caption{DS-Qwen3-8B}
    \label{fig:perturb-curves-gsm8k-prompting-ds-qwen3-8b}
\end{subfigure}
\hfill
\begin{subfigure}[b]{0.48\textwidth}
    \centering
    \includegraphics[width=\textwidth]{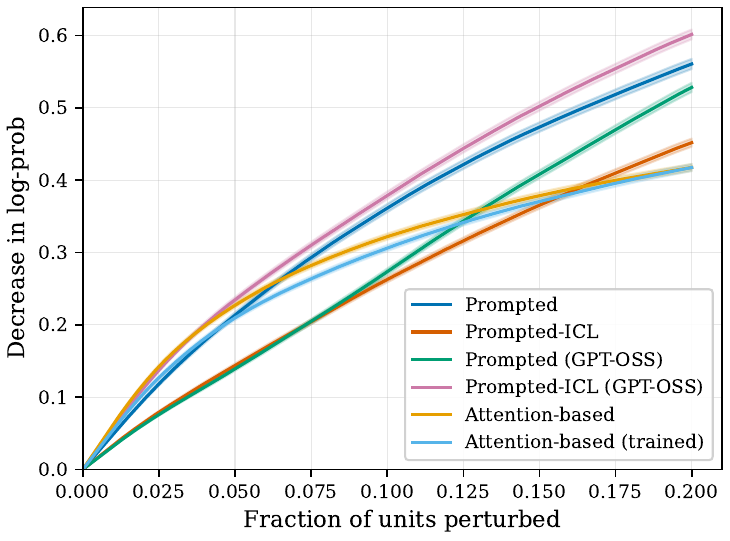}
    \caption{Qwen3-8B}
    \label{fig:perturb-curves-gsm8k-prompting-qwen3-8b}
\end{subfigure}
\caption{Perturbation curves of all prompting baselines and attention-based methods on GSM8K.}
\label{fig:perturb-curves-gsm8k-prompting}
\end{figure}

\begin{figure}[t]
\centering
\begin{subfigure}[b]{0.48\textwidth}
    \centering
    \includegraphics[width=\textwidth]{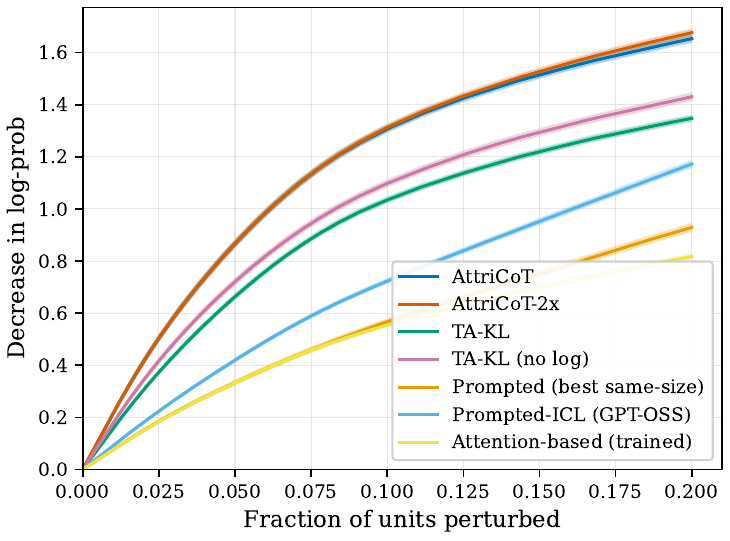}
    \caption{DS-Llama-8B}
    \label{fig:perturb-curves-gsm8k-top-ds-llama-8b}
\end{subfigure}
\hfill
\begin{subfigure}[b]{0.48\textwidth}
    \centering
    \includegraphics[width=\textwidth]{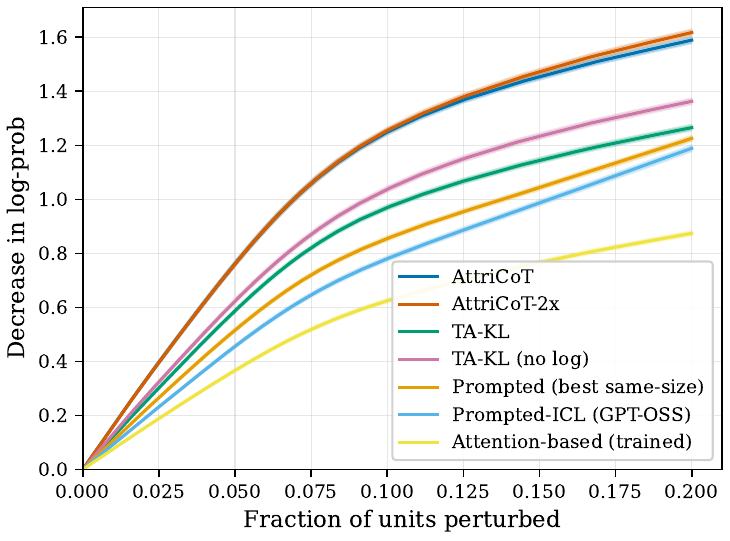}
    \caption{DS-Qwen-14B}
    \label{fig:perturb-curves-gsm8k-top-ds-qwen-14b}
\end{subfigure}

\vspace{0.5em}

\begin{subfigure}[b]{0.48\textwidth}
    \centering
    \includegraphics[width=\textwidth]{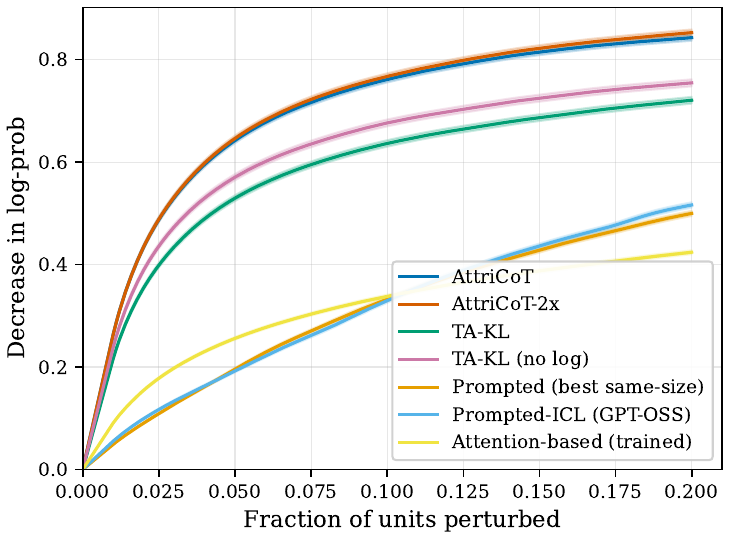}
    \caption{DS-Qwen3-8B}
    \label{fig:perturb-curves-gsm8k-top-ds-qwen3-8b}
\end{subfigure}
\hfill
\begin{subfigure}[b]{0.48\textwidth}
    \centering
    \includegraphics[width=\textwidth]{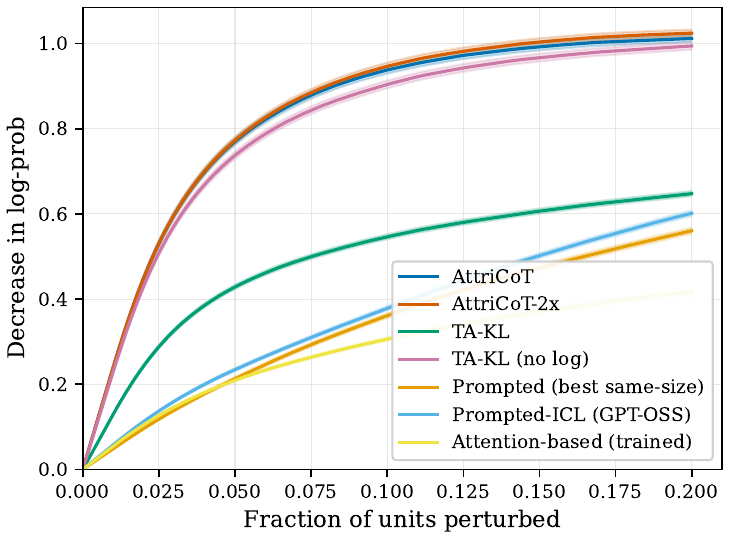}
    \caption{Qwen3-8B}
    \label{fig:perturb-curves-gsm8k-top-qwen3-8b}
\end{subfigure}
\caption{Perturbation curves of all causal attribution methods and best-performing prompting and attention-based methods on GSM8K.}
\label{fig:perturb-curves-gsm8k-top}
\end{figure}

\begin{figure}[t]
\centering
\begin{subfigure}[b]{0.48\textwidth}
    \centering
    \includegraphics[width=\textwidth]{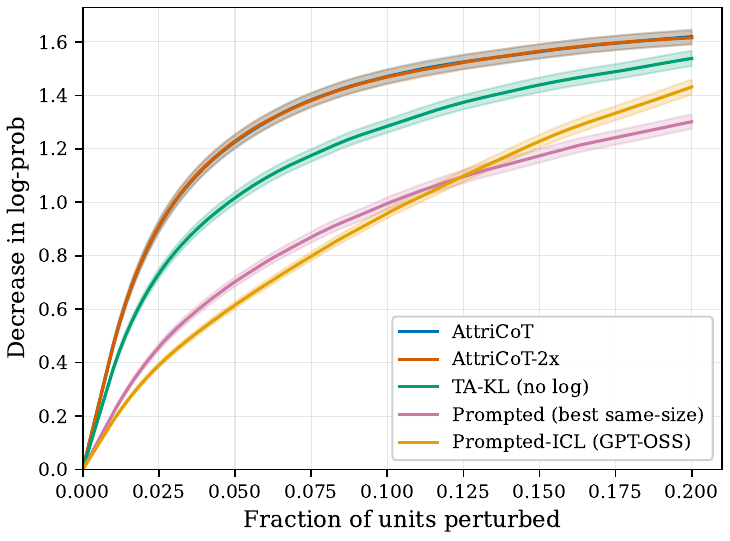}
    \caption{DS-Llama-8B}
    \label{fig:perturb-curves-math500-ds-llama-8b}
\end{subfigure}
\hfill
\begin{subfigure}[b]{0.48\textwidth}
    \centering
    \includegraphics[width=\textwidth]{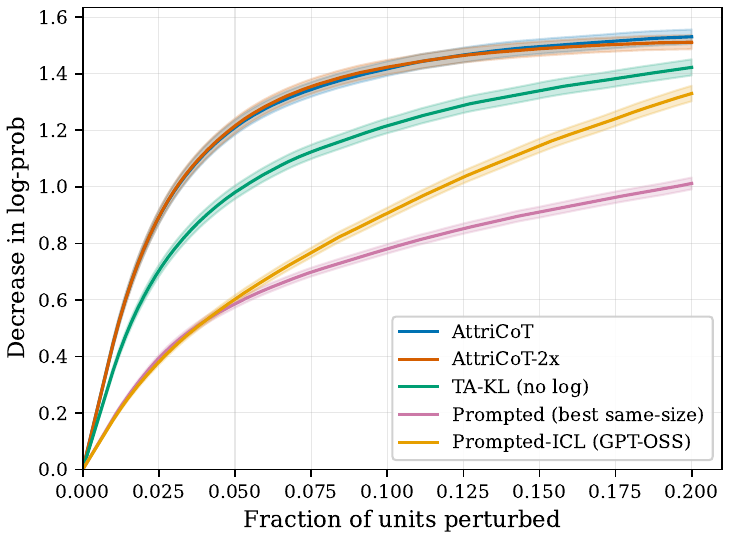}
    \caption{DS-Qwen-14B}
    \label{fig:perturb-curves-math500-ds-qwen-14b}
\end{subfigure}

\vspace{0.5em}

\begin{subfigure}[b]{0.48\textwidth}
    \centering
    \includegraphics[width=\textwidth]{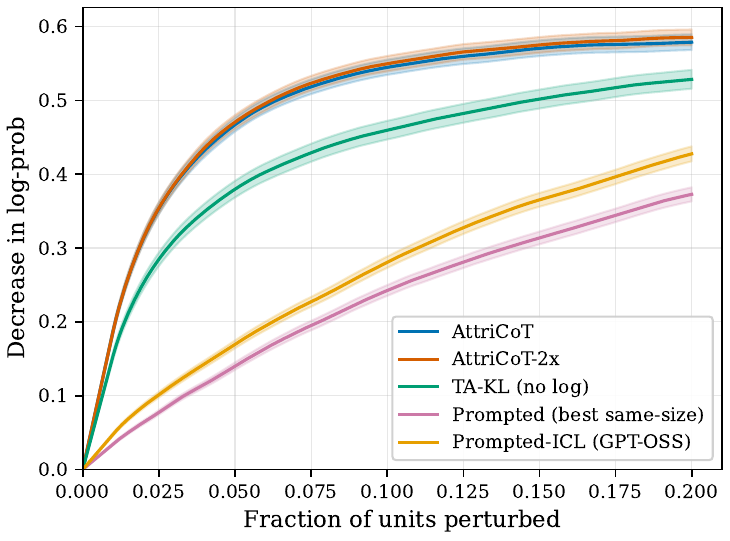}
    \caption{DS-Qwen3-8B}
    \label{fig:perturb-curves-math500-ds-qwen3-8b}
\end{subfigure}
\hfill
\begin{subfigure}[b]{0.48\textwidth}
    \centering
    \includegraphics[width=\textwidth]{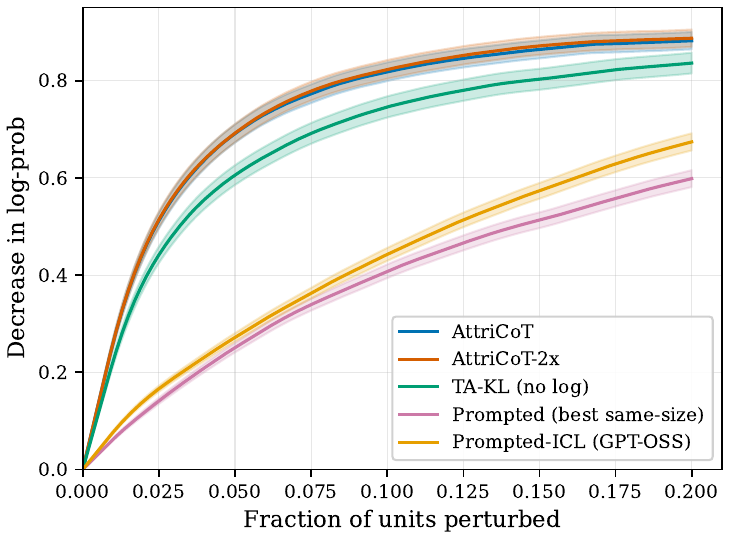}
    \caption{Qwen3-8B}
    \label{fig:perturb-curves-math500-qwen3-8b}
\end{subfigure}
\caption{Perturbation curves on MATH500.}
\label{fig:perturb-curves-math500}
\end{figure}

\begin{figure}[t]
\centering
\begin{subfigure}[b]{0.48\textwidth}
    \centering
    \includegraphics[width=\textwidth]{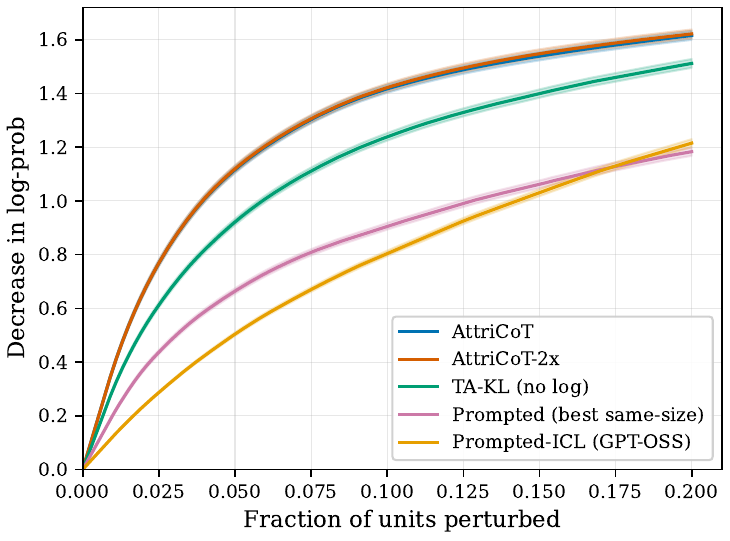}
    \caption{DS-Llama-8B}
    \label{fig:perturb-curves-mmlu-ds-llama-8b}
\end{subfigure}
\hfill
\begin{subfigure}[b]{0.48\textwidth}
    \centering
    \includegraphics[width=\textwidth]{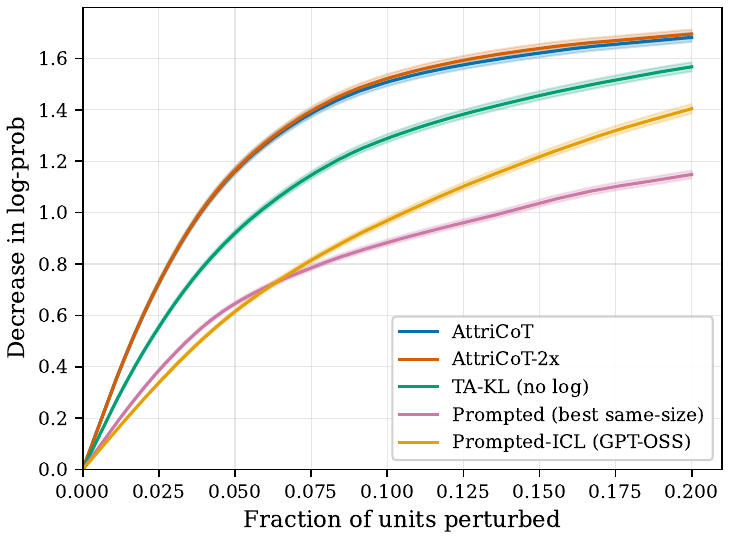}
    \caption{DS-Qwen-14B}
    \label{fig:perturb-curves-mmlu-ds-qwen-14b}
\end{subfigure}

\vspace{0.5em}

\begin{subfigure}[b]{0.48\textwidth}
    \centering
    \includegraphics[width=\textwidth]{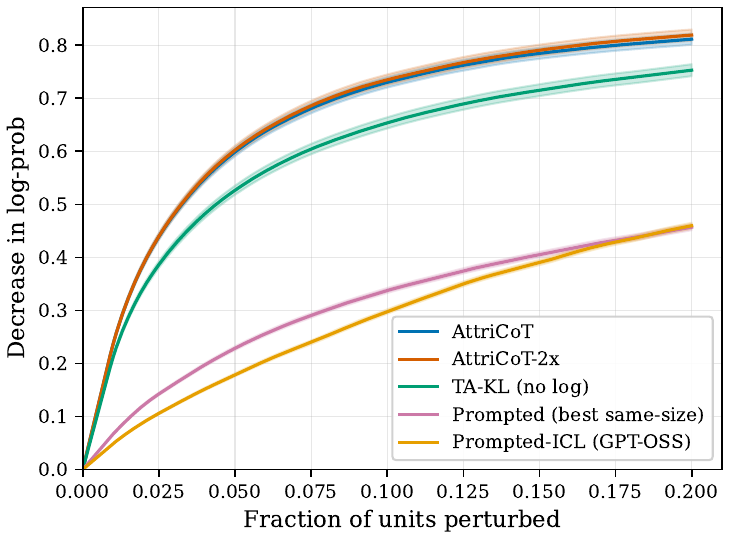}
    \caption{DS-Qwen3-8B}
    \label{fig:perturb-curves-mmlu-ds-qwen3-8b}
\end{subfigure}
\hfill
\begin{subfigure}[b]{0.48\textwidth}
    \centering
    \includegraphics[width=\textwidth]{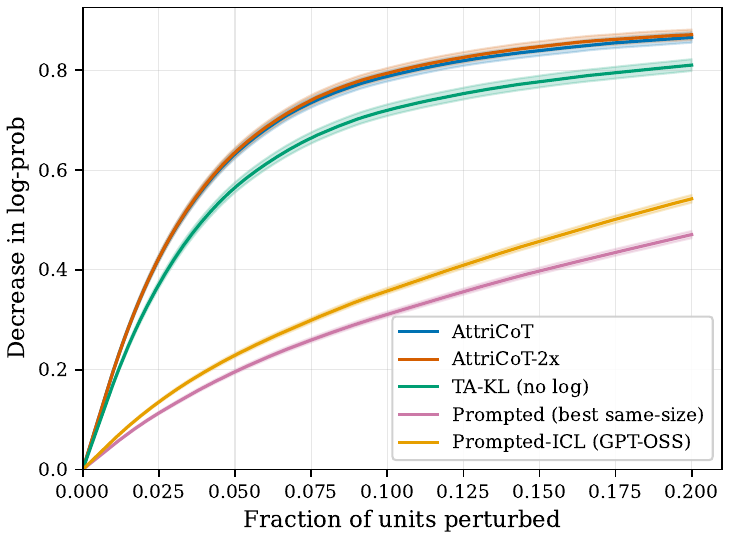}
    \caption{Qwen3-8B}
    \label{fig:perturb-curves-mmlu-qwen3-8b}
\end{subfigure}
\caption{Perturbation curves on MMLU-Pro.}
\label{fig:perturb-curves-mmlu}
\end{figure}

\begin{figure}[t]
\centering
\begin{subfigure}[b]{0.48\textwidth}
    \centering
    \includegraphics[width=\textwidth]{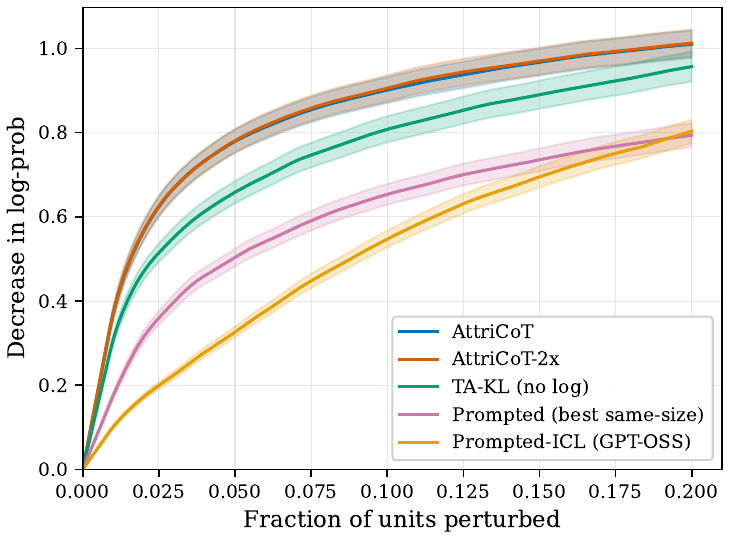}
    \caption{DS-Llama-8B}
    \label{fig:perturb-curves-gpqa-ds-llama-8b}
\end{subfigure}
\hfill
\begin{subfigure}[b]{0.48\textwidth}
    \centering
    \includegraphics[width=\textwidth]{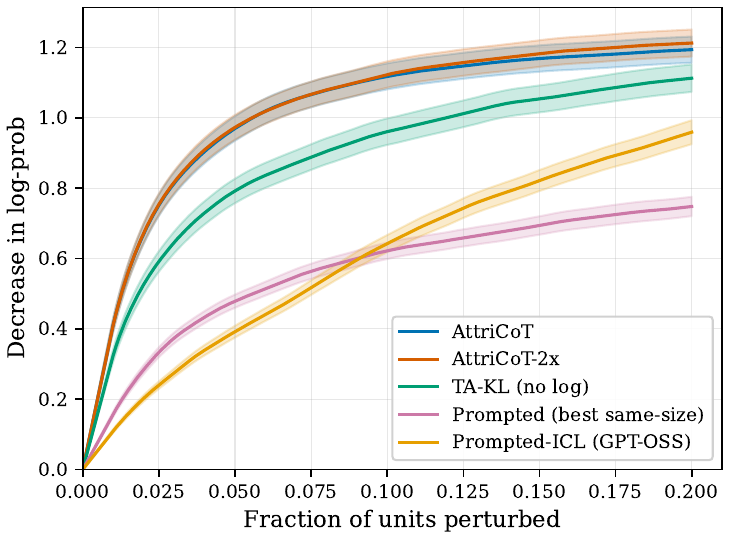}
    \caption{DS-Qwen-14B}
    \label{fig:perturb-curves-gpqa-ds-qwen-14b}
\end{subfigure}

\vspace{0.5em}

\begin{subfigure}[b]{0.48\textwidth}
    \centering
    \includegraphics[width=\textwidth]{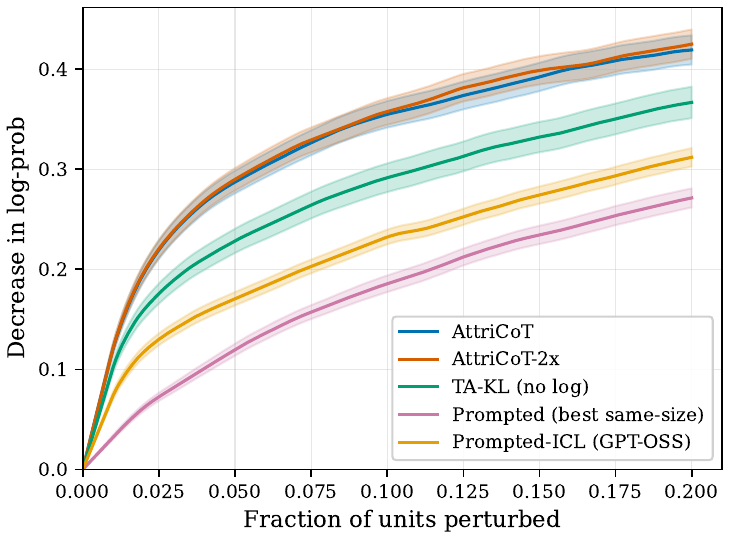}
    \caption{DS-Qwen3-8B}
    \label{fig:perturb-curves-gpqa-ds-qwen3-8b}
\end{subfigure}
\hfill
\begin{subfigure}[b]{0.48\textwidth}
    \centering
    \includegraphics[width=\textwidth]{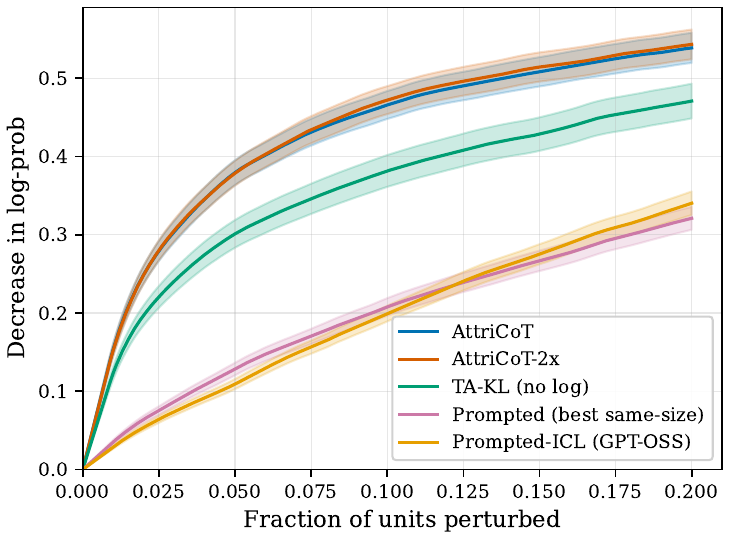}
    \caption{Qwen3-8B}
    \label{fig:perturb-curves-gpqa-qwen3-8b}
\end{subfigure}
\caption{Perturbation curves on GPQA.}
\label{fig:perturb-curves-gpqa}
\end{figure}

\begin{figure}[t]
\centering
\begin{subfigure}[b]{0.48\textwidth}
    \centering
    \includegraphics[width=\textwidth]{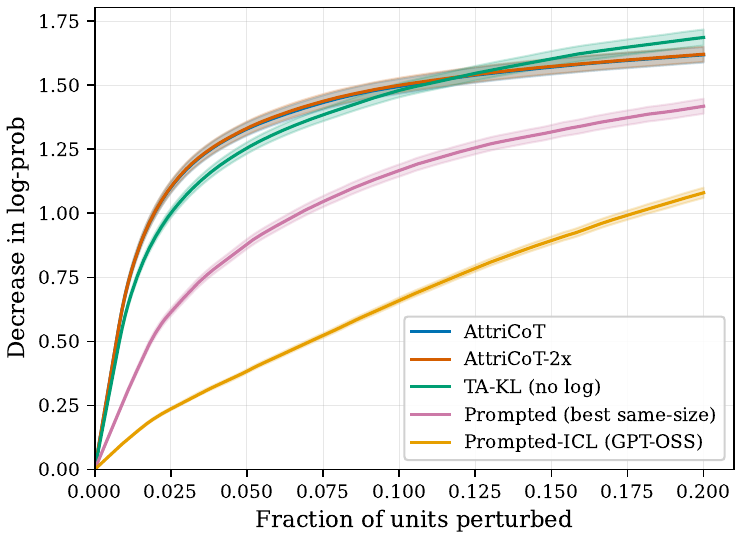}
    \caption{DS-Llama-8B}
    \label{fig:perturb-curves-zebra-ds-llama-8b}
\end{subfigure}
\hfill
\begin{subfigure}[b]{0.48\textwidth}
    \centering
    \includegraphics[width=\textwidth]{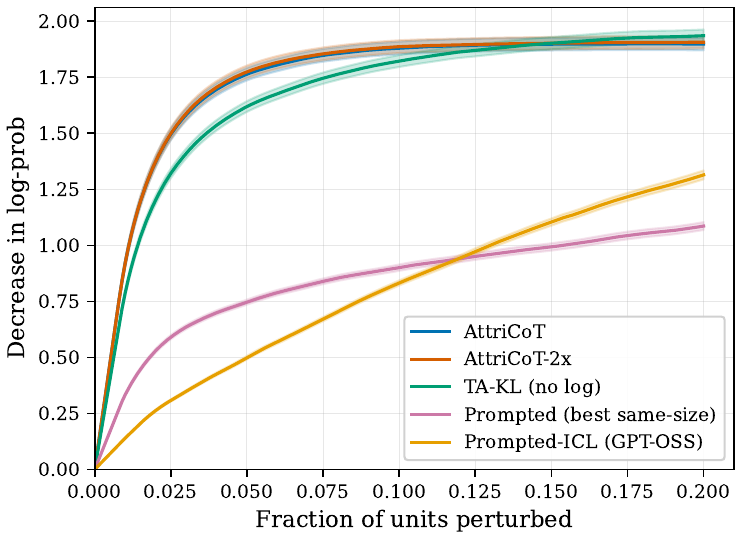}
    \caption{DS-Qwen-14B}
    \label{fig:perturb-curves-zebra-ds-qwen-14b}
\end{subfigure}

\vspace{0.5em}

\begin{subfigure}[b]{0.48\textwidth}
    \centering
    \includegraphics[width=\textwidth]{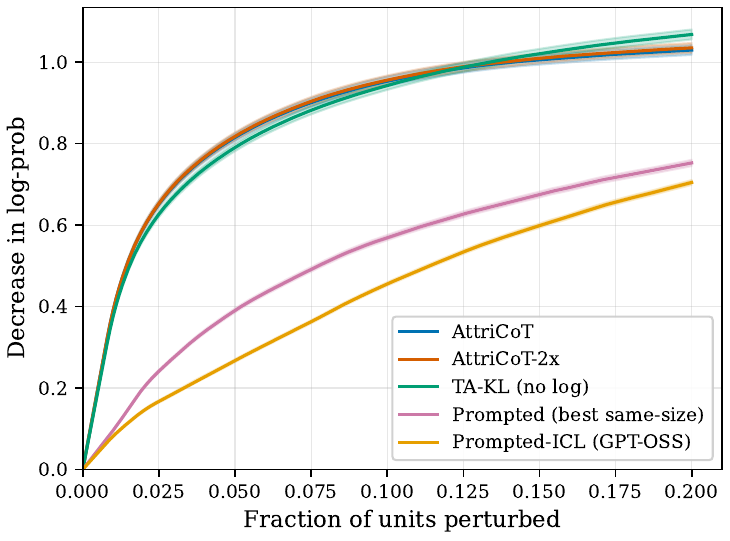}
    \caption{DS-Qwen3-8B}
    \label{fig:perturb-curves-zebra-ds-qwen3-8b}
\end{subfigure}
\hfill
\begin{subfigure}[b]{0.48\textwidth}
    \centering
    \includegraphics[width=\textwidth]{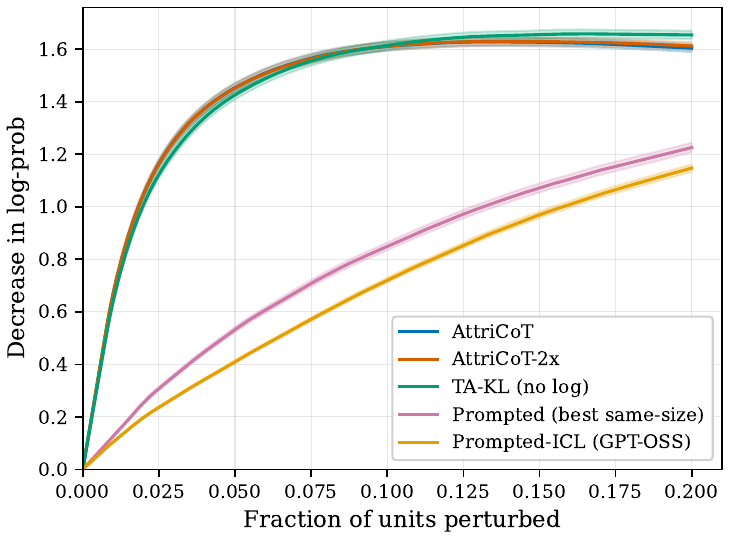}
    \caption{Qwen3-8B}
    \label{fig:perturb-curves-zebra-qwen3-8b}
\end{subfigure}
\caption{Perturbation curves on ZebraLogic.}
\label{fig:perturb-curves-zebra}
\end{figure}

Figures~\ref{fig:perturb-curves-gsm8k-prompting}--\ref{fig:perturb-curves-zebra} plot the full perturbation curves corresponding to the AUPC results in Tables~\ref{tab:aupc_gsm8k} and \ref{tab:aupc-all}. For the GSM8K dataset, since many attribution algorithms were compared in Table~\ref{tab:aupc_gsm8k}, we split the perturbation curves across Figures~\ref{fig:perturb-curves-gsm8k-prompting} and \ref{fig:perturb-curves-gsm8k-top}. Figure~\ref{fig:perturb-curves-gsm8k-prompting} compares all of the prompting baselines and the attention intervention methods, which are competitive with each other. Figure~\ref{fig:perturb-curves-gsm8k-top} then compares the best-performing prompting methods and trained attention intervention with the causal attribution methods (AttriCoT and Thought Anchors-KL). Figures~\ref{fig:perturb-curves-math500}--\ref{fig:perturb-curves-zebra} compare the top-performing methods included in Table~\ref{tab:aupc-all} on the MATH500, MMLU-Pro, GPQA, and ZebraLogic datasets. 

A better perturbation curve is one that shows a more rapid decrease in log probability of output units as more preceding units are perturbed, i.e., a curve that rises more quickly. The curves in Figures~\ref{fig:perturb-curves-gsm8k-prompting}--\ref{fig:perturb-curves-zebra} thus mirror the AUPC values in Tables~\ref{tab:aupc_gsm8k} and \ref{tab:aupc-all}, confirming that AttriCoT's attributions are consistently the most behaviorally faithful to the CoT model's log probabilities.

\clearpage

\subsection{Analysis of attribution scores}
\label{sec:expt:results_addl:analysis}

In this appendix, we show the full set of results for the metrics discussed in Section~\ref{sec:analysis} for summarizing AttriCoT's attribution score matrices ($\mA$ and $\mB$).

\paragraph{Score density} Table~\ref{tab:density} shows values of the score density metric (average absolute score, \eqref{eq:density}) for different pairs of datasets and CoT models (mean and standard error over the samples in each dataset). In general, it appears that score density varies inversely with the length of the model output in terms of number of units. Densities are highest for the GSM8K dataset, which has the shortest CoTs because of the simplicity of the tasks. Densities are lowest for GPQA and ZebraLogic, which tend to have the longest CoTs, while MATH500 and MMLU-Pro are in the middle. Similarly among models, DS-Qwen-14B has the shortest CoTs and correspondingly highest densities, while DS-Qwen3-8B has the longest CoTs and lowest densities. DS-Llama-8B is second-highest in density and Qwen3-8B is third.

\begin{table}[t]
\centering
\caption{Score density metric (average absolute score) applied to AttriCoT's attributions.}
\label{tab:density}
\begin{tabular}{lcccc}
\toprule
Dataset & DS-Llama-8B & DS-Qwen-14B & DS-Qwen3-8B & Qwen3-8B \\
\midrule
GSM8K & 0.2095 $\pm$ 0.0025 & 0.2256 $\pm$ 0.0020 & 0.0471 $\pm$ 0.0010 & 0.0744 $\pm$ 0.0010 \\
MATH500 & 0.0818 $\pm$ 0.0032 & 0.0903 $\pm$ 0.0031 & 0.0279 $\pm$ 0.0010 & 0.0522 $\pm$ 0.0022 \\
GPQA & 0.0321 $\pm$ 0.0024 & 0.0506 $\pm$ 0.0032 & 0.0147 $\pm$ 0.0007 & 0.0198 $\pm$ 0.0011 \\
MMLU-Pro & 0.0787 $\pm$ 0.0013 & 0.1005 $\pm$ 0.0014 & 0.0345 $\pm$ 0.0007 & 0.0477 $\pm$ 0.0009 \\
ZebraLogic & 0.0456 $\pm$ 0.0019 & 0.0638 $\pm$ 0.0023 & 0.0288 $\pm$ 0.0009 & 0.0525 $\pm$ 0.0015 \\
\bottomrule
\end{tabular}
\end{table}

\begin{figure}[t]
\centering
\begin{subfigure}[b]{0.48\textwidth}
    \centering
    \includegraphics[width=\textwidth]{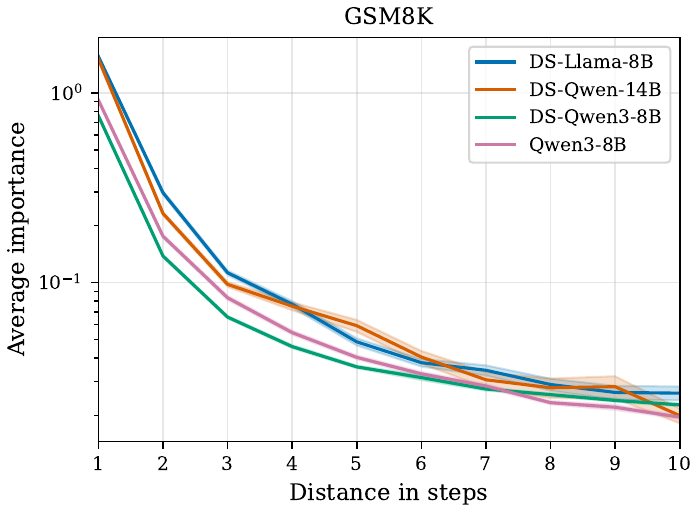}
    \caption{GSM8K}
    \label{fig:locality-abs-all-gsm8k}
\end{subfigure}
\hfill
\begin{subfigure}[b]{0.48\textwidth}
    \centering
    \includegraphics[width=\textwidth]{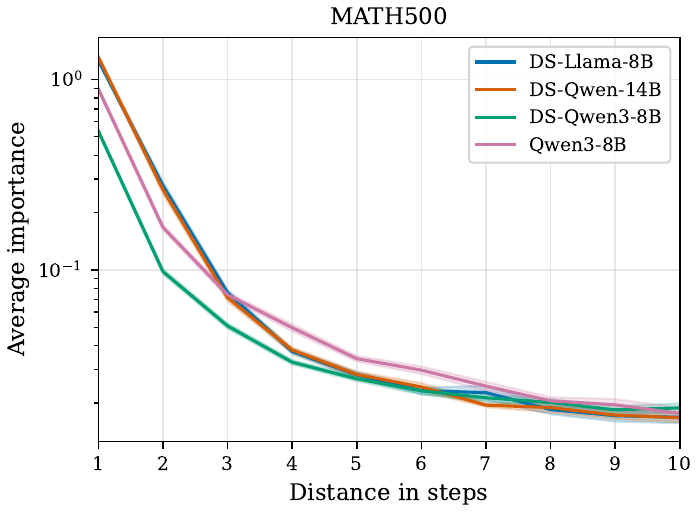}
    \caption{MATH500}
    \label{fig:locality-abs-all-math500}
\end{subfigure}

\vspace{0.5em}

\begin{subfigure}[b]{0.48\textwidth}
    \centering
    \includegraphics[width=\textwidth]{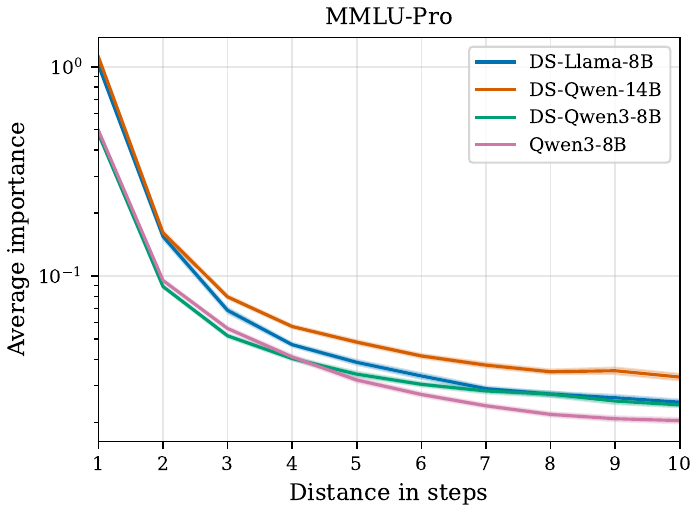}
    \caption{MMLU-Pro}
    \label{fig:locality-abs-all-mmlu}
\end{subfigure}
\hfill
\begin{subfigure}[b]{0.48\textwidth}
    \centering
    \includegraphics[width=\textwidth]{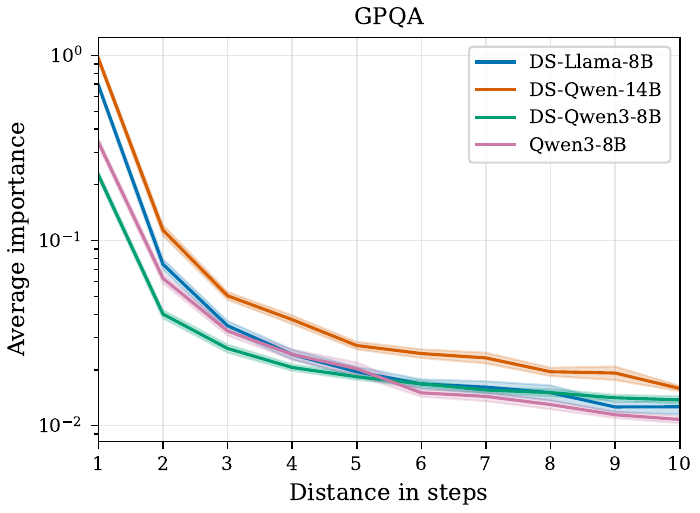}
    \caption{GPQA}
    \label{fig:locality-abs-all-gpqa}
\end{subfigure}

\vspace{0.5em}

\begin{subfigure}[b]{0.48\textwidth}
    \centering
    \includegraphics[width=\textwidth]{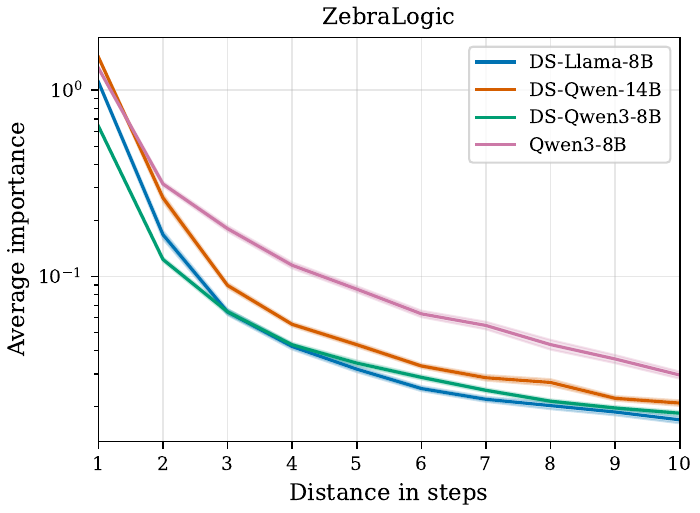}
    \caption{ZebraLogic}
    \label{fig:locality-abs-all-zebra}
\end{subfigure}
\caption{Average importance of CoT units to later CoT units, estimated by AttriCoT, as a function of the distance in steps between the units.}
\label{fig:locality-abs-all}
\end{figure}

\paragraph{Importance by distance} Figure~\ref{fig:locality-abs-all} plots the importance of CoT units to later CoT units (measured by mean absolute score) as a function of the number of steps between them (i.e., absolute distance, see \eqref{eq:locality}). Each curve represents the mean over the dataset samples and the shading indicates one standard error above and below the mean.

Generally, the decay in importance is rapid in the first few steps and then flattens to a long tail. The rates of decay are similar for the four models. The curves for DS-Qwen3-8B and Qwen3-8B tend to be lower than for the other two, in accordance with their generally lower attribution scores seen in Table~\ref{tab:density}. Qwen3-8B on the ZebraLogic dataset (Figure~\ref{fig:locality-abs-all-zebra}) is an exception where its decay is noticeably slower than for the other models.

\begin{figure}[t]
\centering
\begin{subfigure}[b]{0.48\textwidth}
    \centering
    \includegraphics[width=\textwidth]{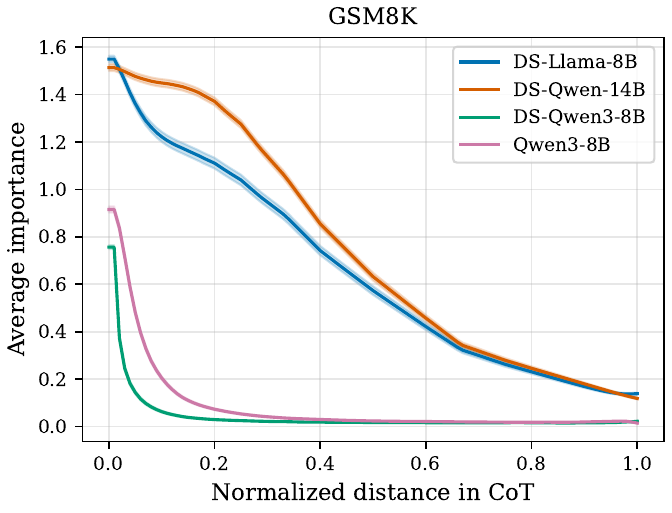}
    \caption{GSM8K}
    \label{fig:locality-norm-all-gsm8k}
\end{subfigure}
\hfill
\begin{subfigure}[b]{0.48\textwidth}
    \centering
    \includegraphics[width=\textwidth]{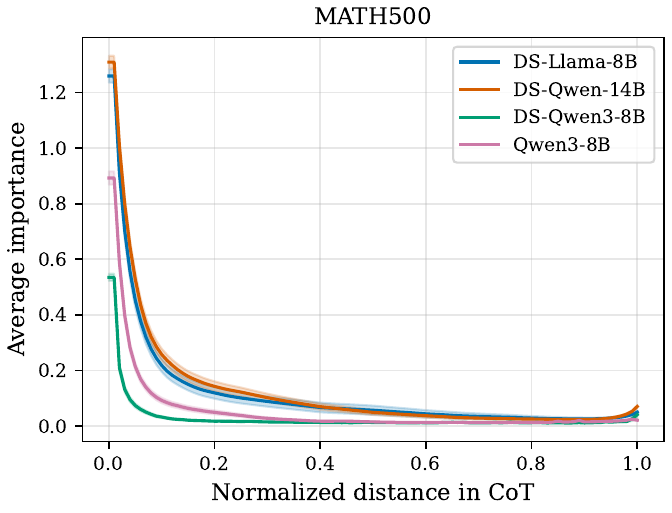}
    \caption{MATH500}
    \label{fig:locality-norm-all-math500}
\end{subfigure}

\vspace{0.5em}

\begin{subfigure}[b]{0.48\textwidth}
    \centering
    \includegraphics[width=\textwidth]{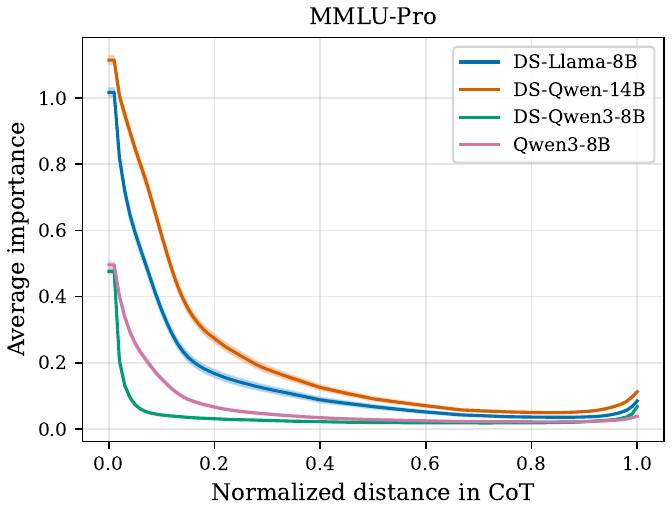}
    \caption{MMLU-Pro}
    \label{fig:locality-norm-all-mmlu}
\end{subfigure}
\hfill
\begin{subfigure}[b]{0.48\textwidth}
    \centering
    \includegraphics[width=\textwidth]{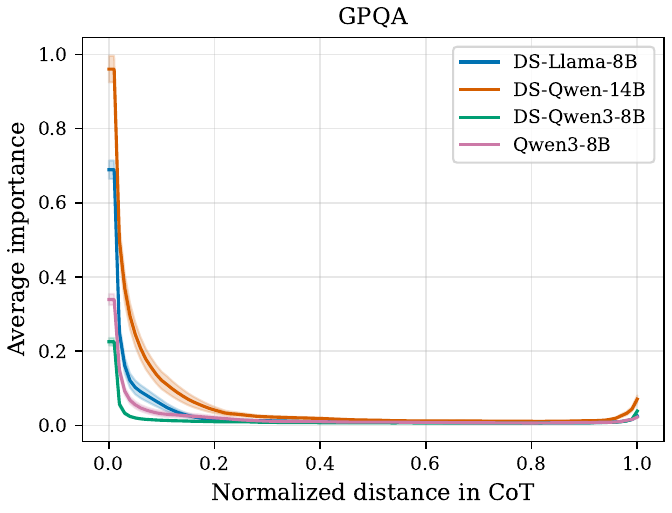}
    \caption{GPQA}
    \label{fig:locality-norm-all-gpqa}
\end{subfigure}

\vspace{0.5em}

\begin{subfigure}[b]{0.48\textwidth}
    \centering
    \includegraphics[width=\textwidth]{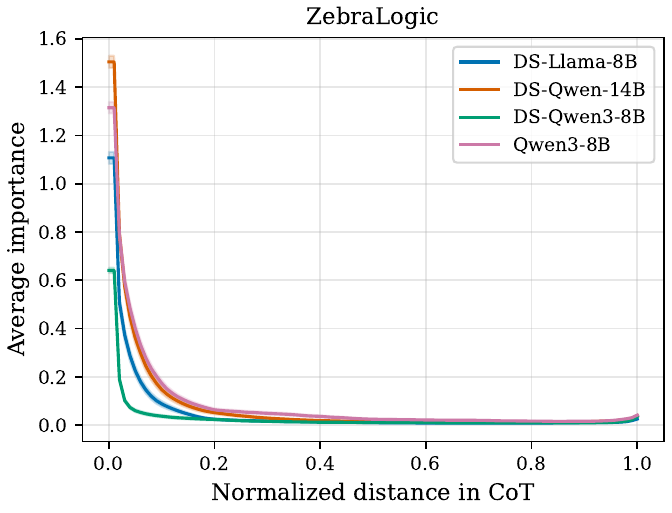}
    \caption{ZebraLogic}
    \label{fig:locality-norm-all-zebra}
\end{subfigure}
\caption{Average importance of CoT units to later CoT units, estimated by AttriCoT, as a function of the normalized distance between the units.}
\label{fig:locality-norm-all}
\end{figure}

Figure~\ref{fig:locality-norm-all} shows an alternative view of importance by distance, where the distance has been normalized to a fraction $d / (T_{\text{CoT}} - 1)$ of the largest possible distance $T_{\text{CoT}} - 1$. To obtain the curves in the figure, the importance-by-distance values for each dataset sample are first interpolated onto a common grid of normalized distances before averaging over samples. 

\clearpage

\begin{figure}[t]
\centering
\begin{subfigure}[b]{0.48\textwidth}
    \centering
    \includegraphics[width=\textwidth]{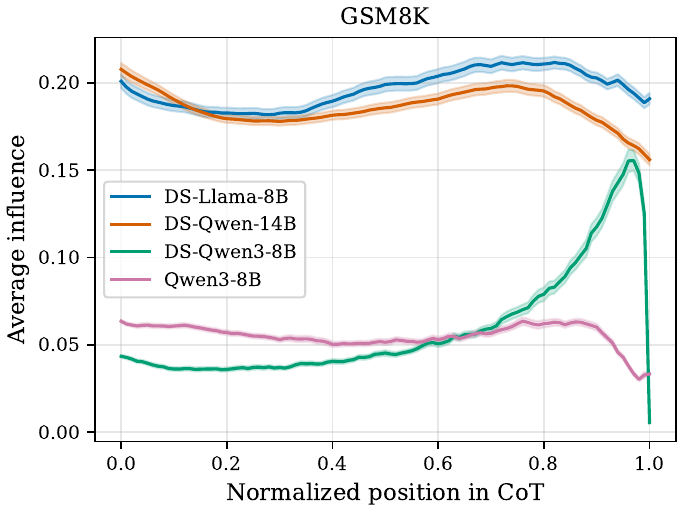}
    \caption{GSM8K}
    \label{fig:influence-all-gsm8k}
\end{subfigure}
\hfill
\begin{subfigure}[b]{0.48\textwidth}
    \centering
    \includegraphics[width=\textwidth]{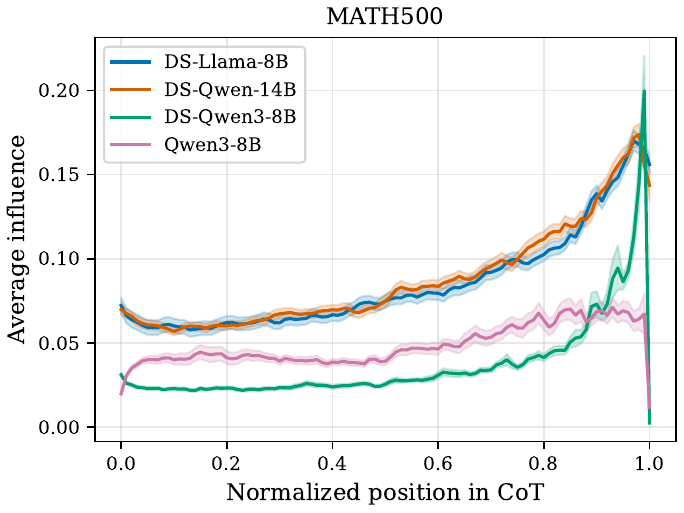}
    \caption{MATH500}
    \label{fig:influence-all-math500}
\end{subfigure}

\vspace{0.5em}

\begin{subfigure}[b]{0.48\textwidth}
    \centering
    \includegraphics[width=\textwidth]{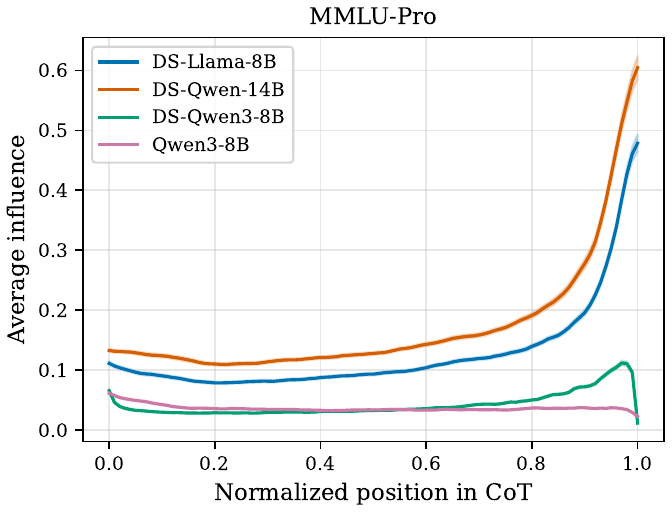}
    \caption{MMLU-Pro}
    \label{fig:influence-all-mmlu}
\end{subfigure}
\hfill
\begin{subfigure}[b]{0.48\textwidth}
    \centering
    \includegraphics[width=\textwidth]{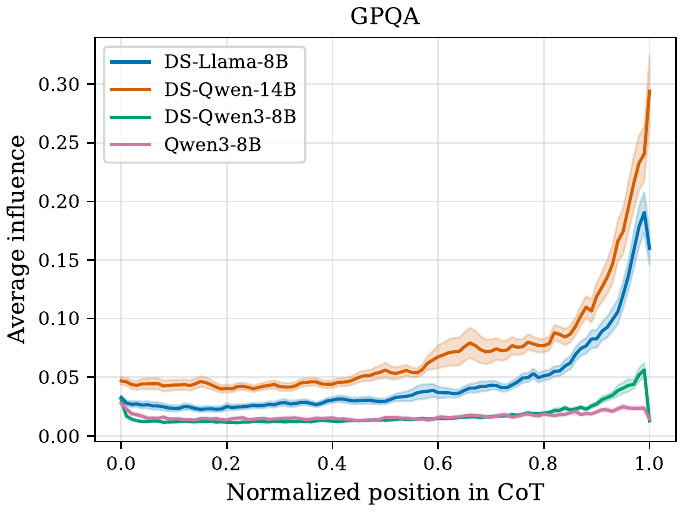}
    \caption{GPQA}
    \label{fig:influence-all-gpqa}
\end{subfigure}

\vspace{0.5em}

\begin{subfigure}[b]{0.48\textwidth}
    \centering
    \includegraphics[width=\textwidth]{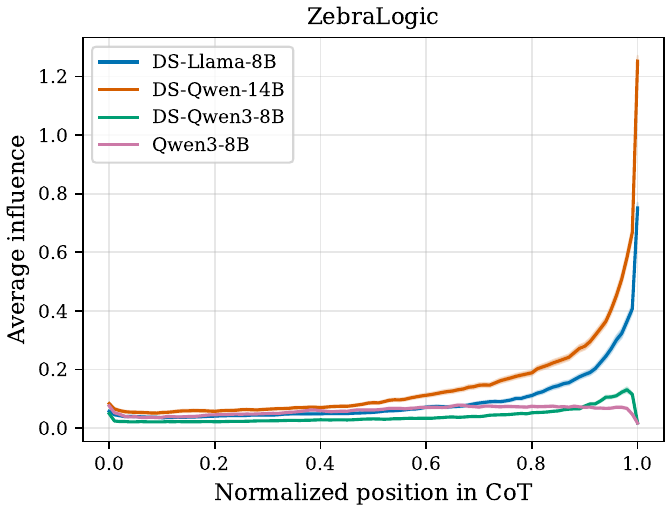}
    \caption{ZebraLogic}
    \label{fig:influence-all-zebra}
\end{subfigure}
\caption{Average influence of CoT units on subsequent output units, estimated by AttriCoT, as a function of normalized position in the CoT.}
\label{fig:influence-all}
\end{figure}

\paragraph{Unweighted average influence} In Figure~\ref{fig:influence-all}, we plot the average influence of CoT units on subsequent output units (now including final answer units), where we take the unweighted average over these subsequent units \eqref{eq:influence}. The curves are plotted as a function of normalized CoT position $\tau / T_{\text{CoT}} \in [0, 1]$ in a similar manner as in Figure~\ref{fig:locality-norm-all}, i.e., by interpolating onto a common grid of positions, averaging, and computing standard errors. Many curves in Figure~\ref{fig:influence-all} (DS-Llama-8B and DS-Qwen-14B on all datasets except GSM8K, and DS-Qwen3-8B on all datasets) exhibit a steep rise as the normalized position approaches $1.0$. We believe that these steep rises are an artifact of taking the unweighted average. In \eqref{eq:influence}, as $\tau$ decreases (going from right to left in the plots), the denominator $T - \tau$ increases linearly. On the other hand, due to the decay of attribution scores with distance seen in Figure~\ref{fig:locality-abs-all}, the sum of absolute scores in \eqref{eq:influence} quickly saturates. The combination of these two effects accounts for the $1 / (T - \tau)$-like behavior in Figure~\ref{fig:influence-all}.

\paragraph{Exponentially-weighted average influence} To address the apparent artifact in Figure~\ref{fig:influence-all}, we also compute exponentially-weighted average influence \eqref{eq:influence-exp} with decay rate $r \in (0, 1)$. The exponential weighting addresses the growing denominator problem in \eqref{eq:influence} by mostly confining the average to units within a \emph{characteristic distance}, proportional to $1 / \log(1/r)$. At the same time, exponential weighting does not completely exclude more distant output units like a sliding window would. We use this interpretation of characteristic distance to set $r$. Specifically, we note that the smallest number of downstream units occurs at $\tau = T_{\text{CoT}}$, where it equals the number of final answer units. Thus, for each dataset-model pair, we compute the average number $\bar{T}_{\text{FA}}$ of final answer units. We then set $r$ so that $90\%$ of the sum of the exponential weights occurs within the first $\bar{T}_{\text{FA}}$ weights, yielding the equations 
\begin{equation}\label{eq:influence-exp-decay-rate}
    1 - r^{\bar{T}_{\text{FA}}} = 0.9 \quad \implies \quad r = (0.1)^{1/\bar{T}_{\text{FA}}}.    
\end{equation}
This setting of $r$ implies that even for the last position $\tau = T_{\text{CoT}}$ in the CoT, most of the mass in the exponential weights falls before the end of the output sequence.

\begin{figure}[t]
\centering
\begin{subfigure}[b]{0.48\textwidth}
    \centering
    \includegraphics[width=\textwidth]{assets/figures/influence_by_position_exp_gsm8k_test.pdf}
    \caption{GSM8K}
    \label{fig:influence-exp-gsm8k}
\end{subfigure}
\hfill
\begin{subfigure}[b]{0.48\textwidth}
    \centering
    \includegraphics[width=\textwidth]{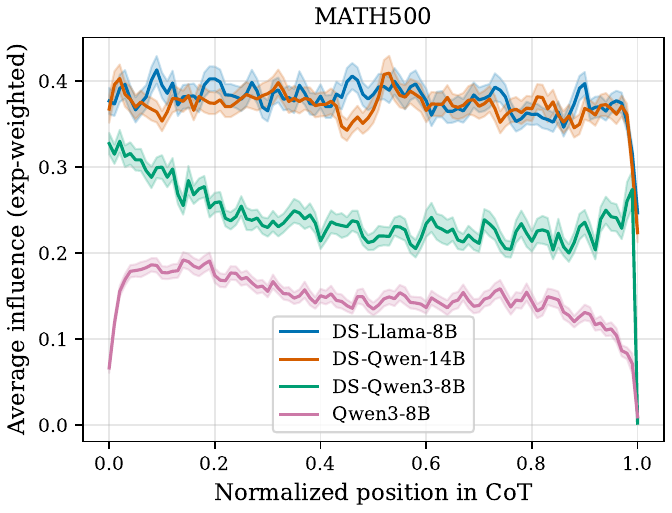}
    \caption{MATH500}
    \label{fig:influence-exp-math500}
\end{subfigure}

\vspace{0.5em}

\begin{subfigure}[b]{0.48\textwidth}
    \centering
    \includegraphics[width=\textwidth]{assets/figures/influence_by_position_exp_mmlu_test.pdf}
    \caption{MMLU-Pro}
    \label{fig:influence-exp-mmlu}
\end{subfigure}
\hfill
\begin{subfigure}[b]{0.48\textwidth}
    \centering
    \includegraphics[width=\textwidth]{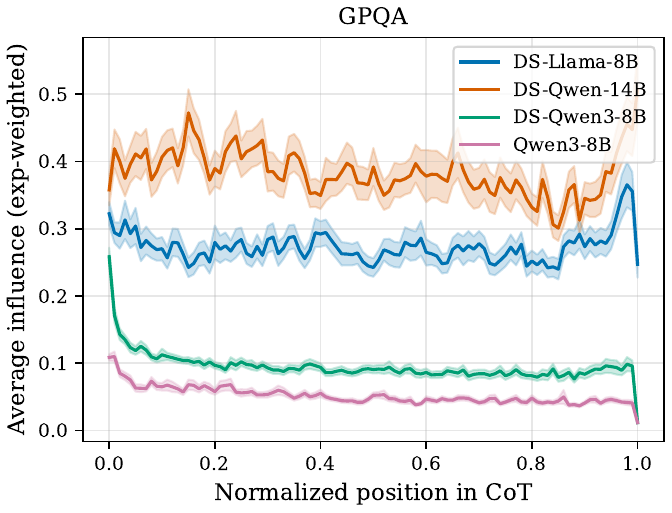}
    \caption{GPQA}
    \label{fig:influence-exp-gpqa}
\end{subfigure}

\vspace{0.5em}

\begin{subfigure}[b]{0.48\textwidth}
    \centering
    \includegraphics[width=\textwidth]{assets/figures/influence_by_position_exp_zebra_logic_pruned.pdf}
    \caption{ZebraLogic}
    \label{fig:influence-exp-zebra}
\end{subfigure}
\caption{Exponentially-weighted average influence of CoT units on subsequent output units, estimated by AttriCoT, as a function of normalized position in the CoT.}
\label{fig:influence-exp}
\end{figure}

Figure~\ref{fig:influence-exp} shows the curves for exponentially-weighted average influence with decay rate determined as described above. The exponential weighting removes the steeply rising artifact seen in Figure~\ref{fig:influence-all}. The middle ranges of the curves are now approximately flat, and the remaining features occur at the beginnings and ends. For DS-Llama-8B and DS-Qwen-14B, exponentially-weighted average influence decreases at the end on the math datasets (GSM8K, MATH500) but increases on the other datasets (MMLU-Pro, GPQA, ZebraLogic). The curves for these two models also decrease at the beginning for some datasets (GSM8K, MMLU-Pro, ZebraLogic) but not others. For DS-Qwen3-8B, exponentially-weighted average influence shows a consistent pattern of decreasing at the beginning and again at the end. Qwen3-8B also shows a decrease at the end, but not a decrease at the beginning on the math datasets.

\clearpage

\begin{figure}[t]
\centering
\begin{subfigure}[b]{0.48\textwidth}
    \centering
    \includegraphics[width=\textwidth]{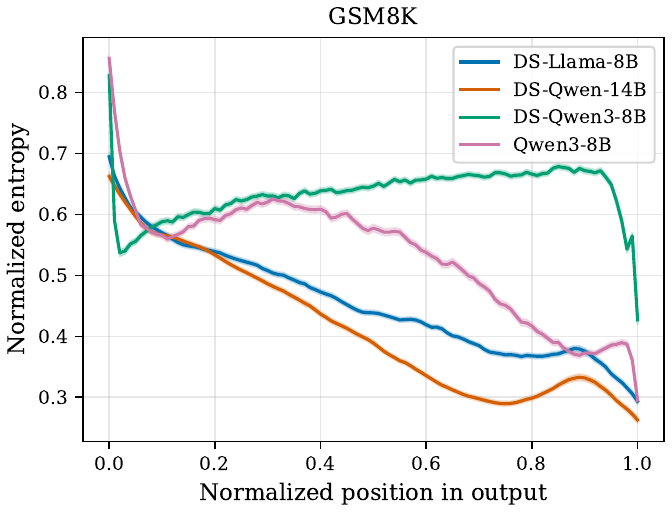}
    \caption{GSM8K}
    \label{fig:entropy-all-gsm8k}
\end{subfigure}
\hfill
\begin{subfigure}[b]{0.48\textwidth}
    \centering
    \includegraphics[width=\textwidth]{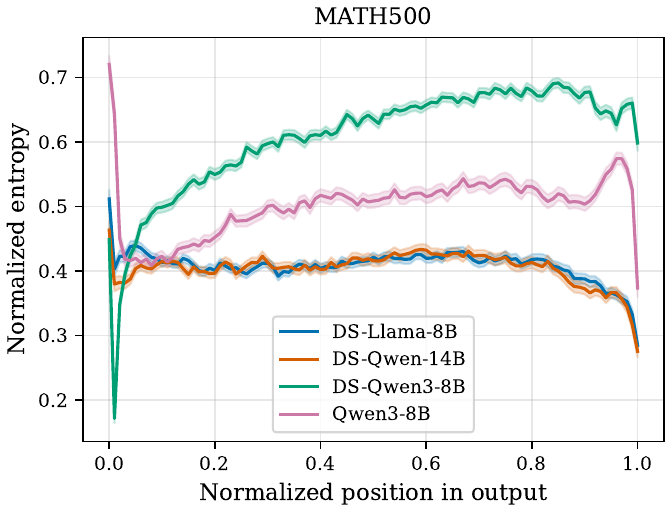}
    \caption{MATH500}
    \label{fig:entropy-all-math500}
\end{subfigure}

\vspace{0.5em}

\begin{subfigure}[b]{0.48\textwidth}
    \centering
    \includegraphics[width=\textwidth]{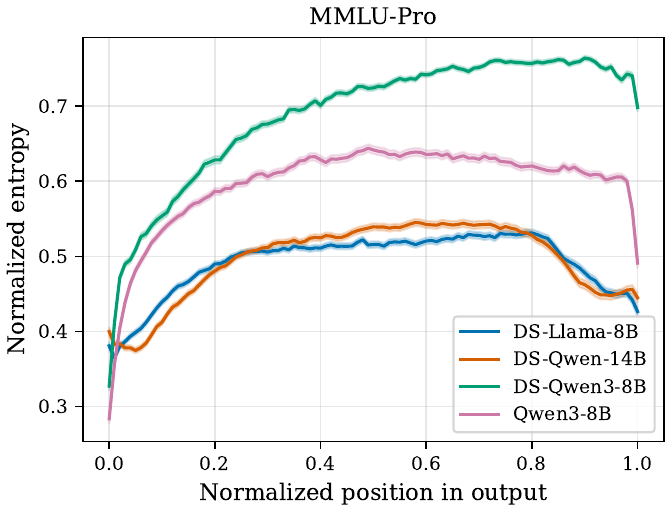}
    \caption{MMLU-Pro}
    \label{fig:entropy-all-mmlu}
\end{subfigure}
\hfill
\begin{subfigure}[b]{0.48\textwidth}
    \centering
    \includegraphics[width=\textwidth]{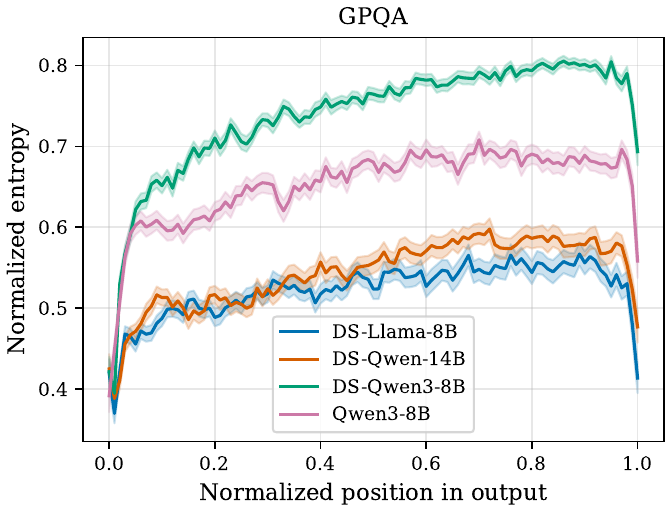}
    \caption{GPQA}
    \label{fig:entropy-all-gpqa}
\end{subfigure}

\vspace{0.5em}

\begin{subfigure}[b]{0.48\textwidth}
    \centering
    \includegraphics[width=\textwidth]{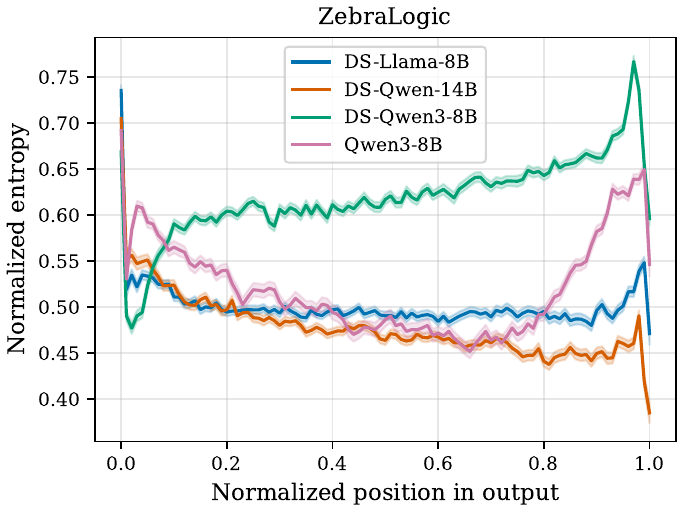}
    \caption{ZebraLogic}
    \label{fig:entropy-all-zebra}
\end{subfigure}
\caption{Normalized entropy of AttriCoT's attribution scores for output units as a function of their position in the output.}
\label{fig:entropy-all}
\end{figure}

\paragraph{Entropy} Figure~\ref{fig:entropy-all} shows the normalized entropy of output units' attribution scores (normalized to the entropy of a uniform distribution) as a function of the output unit's position. Among the four models, DS-Qwen3-8B's attribution scores generally have the highest entropy, followed by Qwen3-8B. This implies that these models' attributions are more diffuse, even after partially correcting for the models' longer CoTs by normalizing the entropy. On the math datasets (GSM8K, MATH500) and ZebraLogic, the entropy is initially higher but quickly drops; this behavior is not present for the general/multiple-domain datasets (MMLU-Pro, GPQA). As the position in the output increases, entropy tends to gradually increase or stay the same (except for three models on GSM8K), before sharply decreasing at the end. For Zebralogic but not the other datasets, entropy builds to a peak before the final decrease. 

\clearpage

\begin{figure}[t]
\centering
\begin{subfigure}[b]{0.48\textwidth}
    \centering
    \includegraphics[width=\textwidth]{assets/figures/input_ratio_gsm8k_test.pdf}
    \caption{GSM8K}
    \label{fig:input-ratio-all-gsm8k}
\end{subfigure}
\hfill
\begin{subfigure}[b]{0.48\textwidth}
    \centering
    \includegraphics[width=\textwidth]{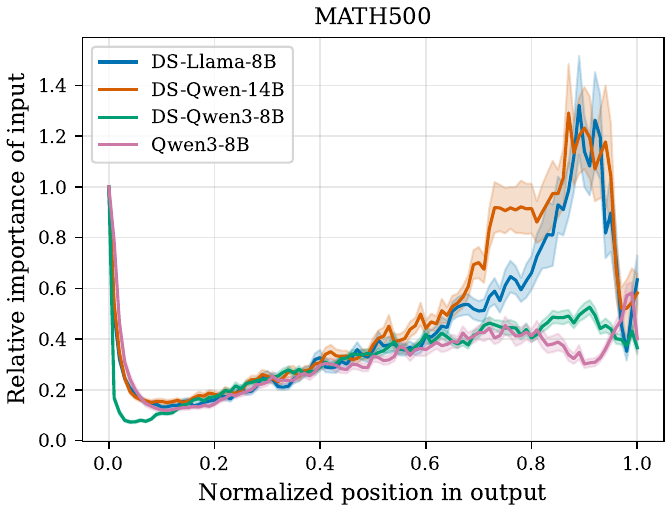}
    \caption{MATH500}
    \label{fig:input-ratio-all-math500}
\end{subfigure}

\vspace{0.5em}

\begin{subfigure}[b]{0.48\textwidth}
    \centering
    \includegraphics[width=\textwidth]{assets/figures/input_ratio_mmlu_test.pdf}
    \caption{MMLU-Pro}
    \label{fig:input-ratio-all-mmlu}
\end{subfigure}
\hfill
\begin{subfigure}[b]{0.48\textwidth}
    \centering
    \includegraphics[width=\textwidth]{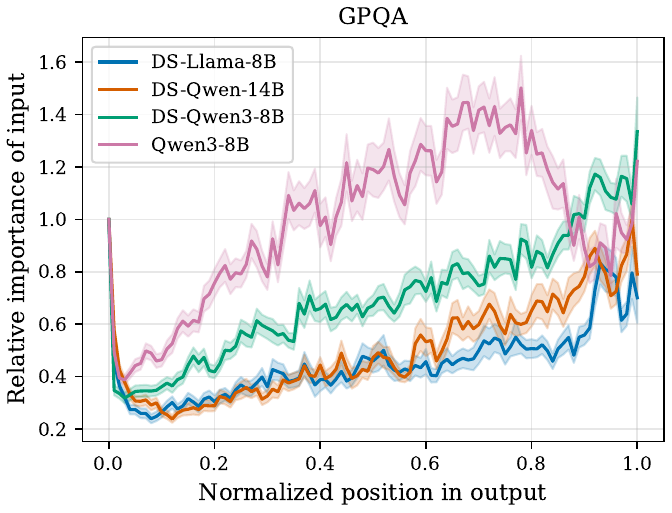}
    \caption{GPQA}
    \label{fig:input-ratio-all-gpqa}
\end{subfigure}

\vspace{0.5em}

\begin{subfigure}[b]{0.48\textwidth}
    \centering
    \includegraphics[width=\textwidth]{assets/figures/input_ratio_zebra_logic_pruned.pdf}
    \caption{ZebraLogic}
    \label{fig:input-ratio-all-zebra}
\end{subfigure}
\caption{Relative importance of input units as a function of normalized position in output.}
\label{fig:input-ratio-all}
\end{figure}

\paragraph{Input ratio.} Figure~\ref{fig:input-ratio-all} plots the input ratio metric for output units, including for the MATH500 and GPQA datasets that were omitted from Figure~\ref{fig:input-ratio-main} in Section~\ref{sec:analysis:results}. The patterns for MATH500 and GPQA are as described in Section~\ref{sec:analysis:results}.

\begin{figure}[t]
\centering
\begin{subfigure}[b]{0.48\textwidth}
    \centering
    \includegraphics[width=\textwidth]{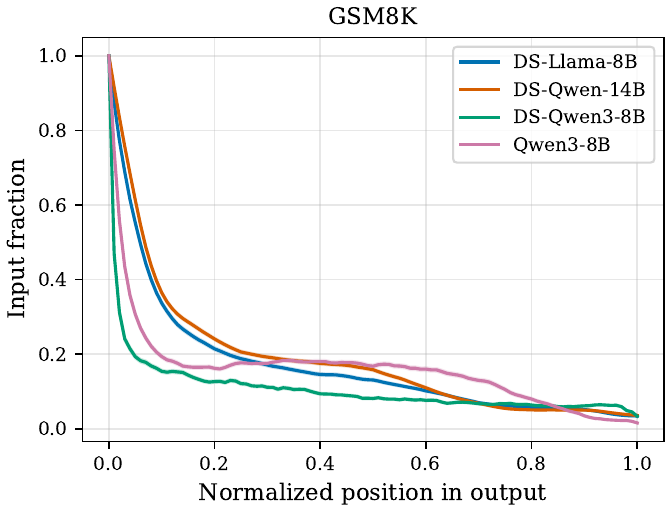}
    \caption{GSM8K}
    \label{fig:input-fraction-all-gsm8k}
\end{subfigure}
\hfill
\begin{subfigure}[b]{0.48\textwidth}
    \centering
    \includegraphics[width=\textwidth]{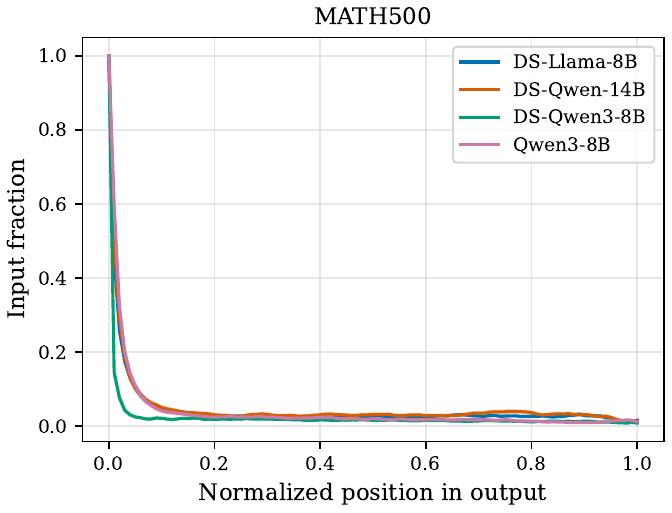}
    \caption{MATH500}
    \label{fig:input-fraction-all-math500}
\end{subfigure}

\vspace{0.5em}

\begin{subfigure}[b]{0.48\textwidth}
    \centering
    \includegraphics[width=\textwidth]{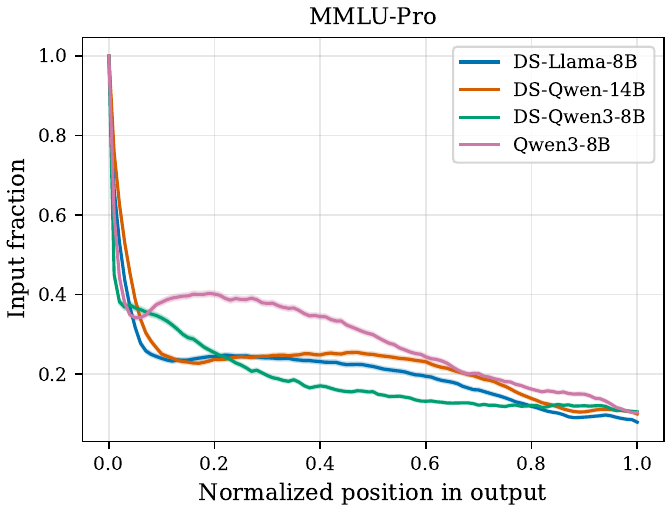}
    \caption{MMLU-Pro}
    \label{fig:input-fraction-all-mmlu}
\end{subfigure}
\hfill
\begin{subfigure}[b]{0.48\textwidth}
    \centering
    \includegraphics[width=\textwidth]{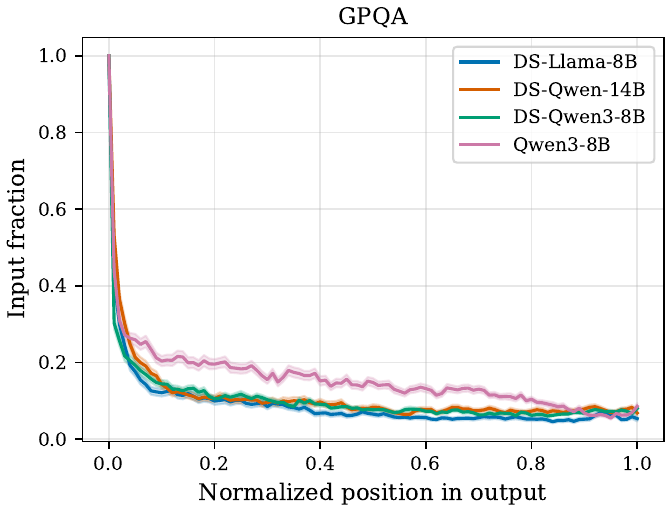}
    \caption{GPQA}
    \label{fig:input-fraction-all-gpqa}
\end{subfigure}

\vspace{0.5em}

\begin{subfigure}[b]{0.48\textwidth}
    \centering
    \includegraphics[width=\textwidth]{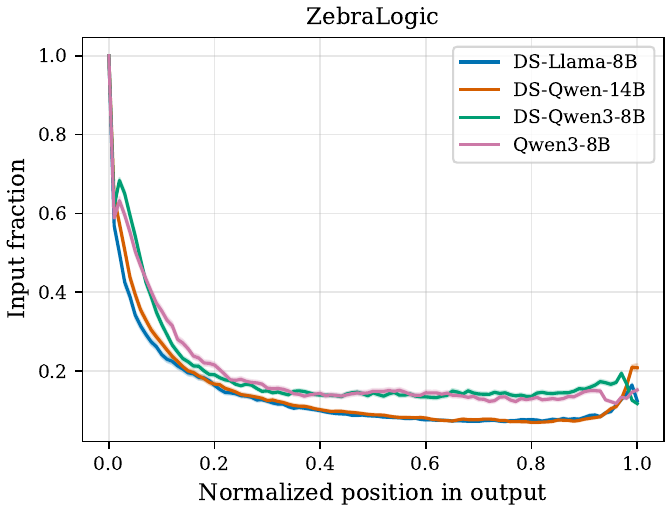}
    \caption{ZebraLogic}
    \label{fig:input-fraction-all-zebra}
\end{subfigure}
\caption{Input fraction by normalized position in output.}
\label{fig:input-fraction-all}
\end{figure}

\paragraph{Input fraction.} We also introduce the Input fraction (see \ref{eq:input-fraction}). For a given output unit, the metric quantify the fraction between the sum of the input units attribution compared to the sum of the attribution scores of all previous output units. It means that if the \emph{Input fraction} of a given output unit is high, the input unit is driving the attribution of the unit of interest relatively to output units (i.e. input units are tend to be more responsible for the attribution of the unit of interest). Figure \ref{fig:input-fraction-all} present the \emph{input fraction} by normalized position of the units in output. Across models and dataset, the Input fraction tends to start high and then gradually decay along the $x$ axis. When generating the initial units, the models rely heavily on the context as they reformulate and appropriate the problem to themselves. Further, models rely more on the past units that they generated. Importantly, for general reasoning and knowledge benchmarks (i.e. MMLU-Pro - Figure \ref{fig:input-fraction-all-mmlu}, GPQA-Diamond - Figure \ref{fig:input-fraction-all-gpqa}, and ZebraLogic - Figure \ref{fig:input-fraction-all-zebra}), the Input fraction decreases less quickly, and remains high relatively to mathematical problems for some models. Non-mathematical datasets includes a list of potential solution in the input prompt. To solve problems, models need to recall the context more frequently to discuss the potential options. This phenomenon seems to align with the Input fraction scores obtained.

\end{document}